\PassOptionsToPackage{table}{xcolor}
\documentclass[10pt,twocolumn,letterpaper]{article}

\usepackage[algorithms]{wacv} 

\definecolor{wacvblue}{rgb}{0.21,0.49,0.74}
\usepackage[breaklinks,colorlinks,allcolors=wacvblue]{hyperref}
\usepackage{epsfig}
\usepackage{multirow}
\usepackage{colortbl} 
\usepackage{placeins}
\usepackage{algorithm}
\usepackage{algorithmic}
\usepackage{float}

\usepackage{subcaption}     
\makeatletter
\@ifundefined{eg}{}{}
\@ifundefined{ie}{}{}
\@ifundefined{etal}{\emph{et al}\onedot}{}
\makeatother

\newcommand{\todo}[1]{\textcolor{red}{\textbf{[TODO]}}}
\newcommand{\note}[1]{}

\def\confName{WACV}
\def\confYear{2027}

\definecolor{spectralbg}{RGB}{240,248,238}
\definecolor{semanticbg}{RGB}{252,240,240}
\definecolor{oursbg}{RGB}{235,247,246}

\newcommand{\venue}[1]{\textcolor{black!45}{\scriptsize (#1)}}

\begin{document}

\title{Don’t Ask the LLM: Visual Token Pruning with Spectral Structure and Semantic Relevance}

\author{
Intaek Shin, \quad
Geuntaek Lim, \quad
Sungjune Park, \quad
Yukyung Choi\setcounter{footnote}{1}\thanks{Corresponding author} \\
Sejong University, South Korea \\
{\small \texttt{\{itshin,ykchoi\}@rcv.sejong.ac.kr}}
}

\maketitle

\begingroup
\renewcommand{\thefootnote}{}
\footnotetext{%
\scriptsize
\copyright\ 2027 IEEE. Personal use of this material is permitted.
Permission from IEEE must be obtained for all other uses,
in any current or future media, including reprinting/republishing
this material for advertising or promotional purposes,
creating new collective works, for resale or redistribution
to servers or lists, or reuse of any copyrighted component
of this work in other works.
Accepted to WACV 2027.%
}
\endgroup

\begin{abstract}
Large vision-language models (LVLMs) represent images and videos as long visual-token sequences, substantially increasing inference costs.
These costs motivate visual-token pruning, whose effectiveness depends on both what signal is used for selection and when that signal becomes available.
Our analysis reveals complementary limitations: visual-side \texttt{[CLS]} attention produces scattered, background-biased selections under aggressive compression, while stronger query-conditioned signals become available after substantial computation within the large language model (LLM).
This raises a natural question: can structurally coherent, query-aware token selection be performed before consulting the LLM?
We therefore propose \textbf{ST-Pruner} (\textbf{S}pectral-\textbf{T}extual \textbf{Pruner}), a training-free framework for pre-LLM visual-token selection.
It combines attention-independent spectral saliency with inverted patch--text similarity through parameter-free multiplicative fusion.
Across diverse LVLMs, ST-Pruner retains over $95\%$ of full-token performance using at most $11.1\%$ of visual tokens and achieves a $3.2\times$ end-to-end speedup on LLaVA-NeXT. Code is
available at \url{https://github.com/sejong-rcv/ST-Pruner}.
\end{abstract}

\section{Introduction}
\label{sec:intro}

\begin{figure}[!t]
    \centering
    \includegraphics[
        trim={7pt 15pt 7pt 0pt},
        clip,
        width=\columnwidth
    ]{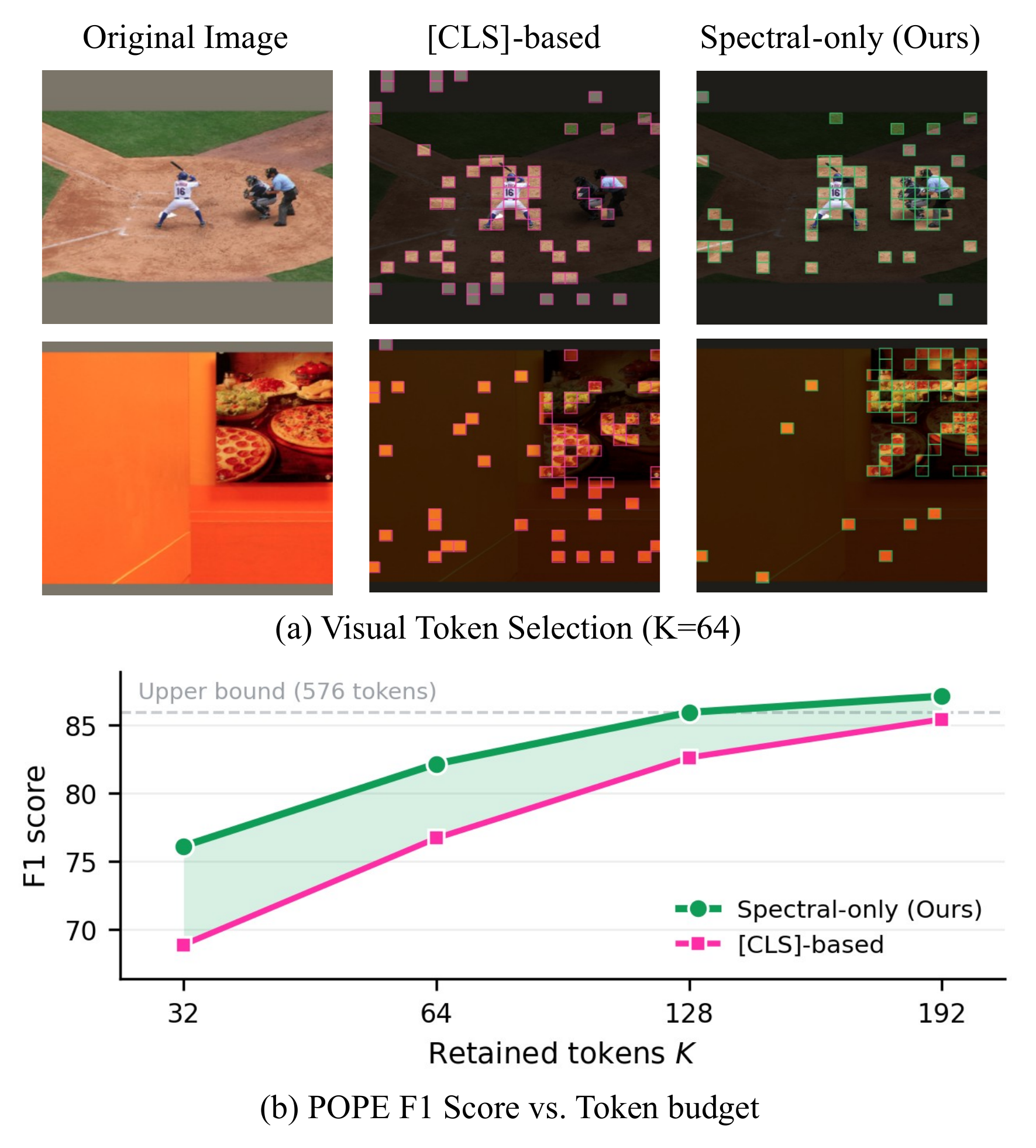}
    \caption{
        \textbf{Structural Saliency Comparison.}
        (a) Selection of $K=64$ tokens using \texttt{[CLS]} attention
        and our spectral criterion.
        (b) F1 score on POPE across token budgets using each structural signal.
    }
    \label{fig:fig1}
\end{figure}

Building on recent advances in large language models (LLMs)~\cite{touvron2023llama,bai2023qwen}, large vision-language models (LVLMs)~\cite{li2023blip,liu2023visual,liu2024llava15, liu2024llavanext,qwen2025vl72b,li2024llava,lin2024video}
extend language-model reasoning to visual inputs represented as visual-token sequences.
As visual-token sequences grow with resolution, multi-view processing, and temporal inputs, quadratic self-attention costs~\cite{NIPS2017_3f5ee243} have motivated extensive work on visual token pruning~\cite{chen2024image,pmlr-v267-zhang25s, Yang_2025_CVPR,Zhang_2025_ICCV,Alvar_2025_CVPR}.

A central challenge in visual token pruning is to remove substantial redundancy while preserving coherent visual structure and query-relevant evidence.
However, commonly used signals for these two criteria exhibit complementary challenges.

\begin{figure*}[!t]
    \centering
    \includegraphics[
    trim={45pt 0pt 5pt 3pt},
    width=\textwidth,  
    ]{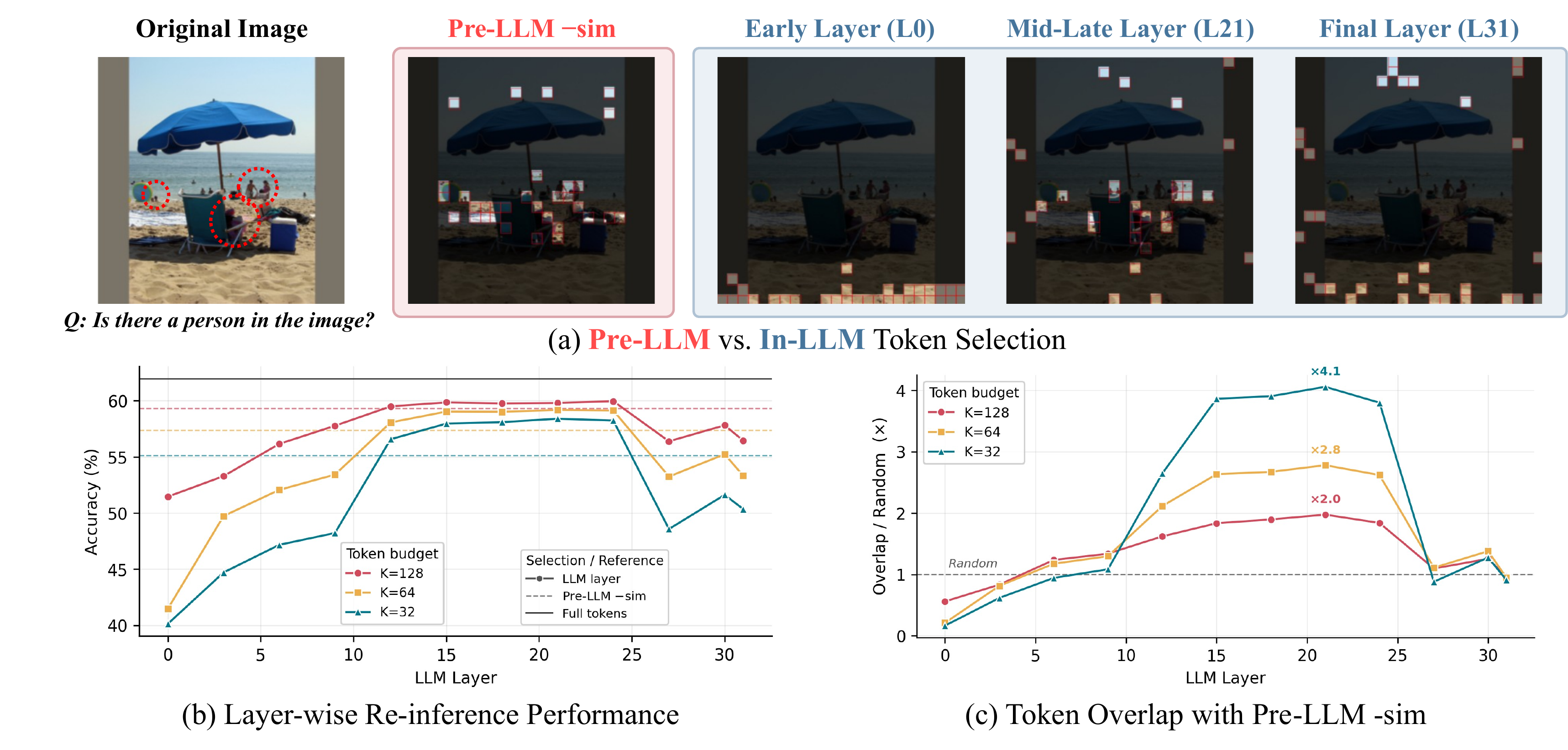}
    \caption{
        \textbf{Why Not Ask the LLM?}
        (a) Pre-LLM inverted patch--text similarity ($-\mathrm{sim}$) and in-LLM query-to-patch attention at representative decoder layers, visualized by their top-$K$ token selections.
        (b--c) Evaluations on GQA ($n=12{,}578$).
        (b) Re-inference accuracy when retaining only the top-$K$
        tokens selected by each LLM layer, with the full-token
        model and pre-LLM $-\mathrm{sim}$ as references.
        (c) Layer-wise overlap between the pre-LLM $-\mathrm{sim}$
        and in-LLM top-$K$ token sets, normalized by the
        expected overlap of two random $K$-token subsets,
        such that $1\times$ denotes random overlap.
    }
    \label{fig:fig2}
\end{figure*}

First, pre-LLM pruning based on visual signals commonly relies on \texttt{[CLS]}-to-patch attention, which ranks patches using a global image-level attention signal~\cite{Zhang_2025_ICCV,Yang_2025_CVPR}.
At low token budgets, however, \texttt{[CLS]}-based selection becomes noisy and background-biased, scattering selections across background regions and missing foreground objects (Figure~\ref{fig:fig1}(a)).

Second, query relevance is often estimated from query-to-patch attention inside the LLM~\cite{chen2024image,pmlr-v267-zhang25s}, but its quality varies substantially across decoder layers.
As illustrated in Figure~\ref{fig:fig2}(a), early-layer selections are strongly biased toward later-positioned visual tokens, consistent with the positional attention shift induced by rotary position embedding (RoPE)~\cite{Zhang_2025_ICCV}.
In the mid-to-late layers, attention instead concentrates on question-relevant evidence before becoming dispersed again near the end of the decoder.
Layer-wise re-inference using the attention-selected top-$K$ tokens from each layer (Figure~\ref{fig:fig2}(b)) shows that selections from the mid-to-late layers yield the strongest downstream performance.

Thus, the two signals expose complementary limitations: visual-side attention struggles to preserve coherent structure under aggressive compression, whereas stronger query-conditioned selections are observed in the mid-to-late layers, after substantial LLM computation.

This raises the central question behind our work: \emph{can both structural coherence and query awareness be obtained before consulting the LLM?}
We therefore propose \textbf{ST-Pruner} (\textbf{S}pectral-\textbf{T}extual \textbf{Pruner}), a training-free pre-LLM framework that combines attention-independent spectral saliency from the low-rank structure of visual features with inverted patch--text similarity through parameter-free multiplicative fusion.

To preserve visual structure without relying on a single global attention distribution, we characterize each patch by its participation in the dominant low-rank structure of the visual feature matrix using singular value decomposition (SVD).
The resulting spectral saliency produces more spatially coherent selections with reduced background scatter.
On POPE, an object-presence benchmark, F1 increases from $76.7$ to $82.1$ at $K=64$ over \texttt{[CLS]}-based selection (Figure~\ref{fig:fig1}(b)).

Spectral saliency alone, however, does not account for the input query.
Prior CLIP~\cite{radford2021learning} interpretability studies report counter-intuitive local image--text responses attributed to pooling-induced \emph{semantic shift}~\cite{li2022exploring}.
Motivated by this observation, we use inverted patch--text similarity as a pre-LLM relevance signal.
As shown in Figure~\ref{fig:fig2}(a,c), the normalized overlap between its top-$K$ set and the in-LLM selections increases in the mid-to-late regime, reaching $4.1\times$ the random-overlap expectation at $K=32$.
This provides a pre-LLM surrogate for the query-aware selections observed in mid-to-late LLM layers without decoder-side scoring.
Together, the two cues provide complementary structural and semantic evidence before LLM inference.

\noindent\textbf{Contributions.}
Our contributions are threefold:
\begin{itemize}

\item We conduct a systematic analysis of commonly used visual-token selection signals and identify complementary limitations: \texttt{[CLS]}-based selection becomes noisy and background-biased under aggressive compression, while stronger query-conditioned selections appear only after substantial LLM computation.

\item Guided by these observations, we propose \textbf{ST-Pruner}, a training-free pre-LLM framework that combines attention-independent spectral saliency from low-rank visual structure with inverted patch--text similarity through parameter-free multiplicative fusion.

\item We demonstrate strong compression--performance trade-offs across diverse image and video LVLMs. ST-Pruner retains over $95\%$ of full-token performance with at most $11.1\%$ of the visual tokens, and achieves a $3.2\times$ end-to-end speedup on LLaVA-NeXT while retaining $99.5\%$ of the full-token POPE F1.

\end{itemize}


\section{Related Work}
\label{sec:related}

\subsection{Training-Free Visual Token Reduction}

Training-free visual token reduction methods can be broadly categorized by where token importance is estimated: inside the LLM or before LLM inference.
In-LLM methods estimate visual-token importance from decoder-side signals and prune tokens at intermediate layers.
FastV~\cite{chen2024image} and SparseVLM~\cite{pmlr-v267-zhang25s}, for example, use text-to-visual attention for token selection.
Because importance is estimated only after entering the decoder, their computational savings depend on the layer at which pruning occurs.
Pre-LLM methods instead reduce visual tokens before decoder execution.
Some rely only on visual-side information, including attention, redundancy, diversity, spatial coverage, approximation error, or token sensitivity~\cite{shang2025llava,Yang_2025_CVPR,Zhang_2025_ICCV, Alvar_2025_CVPR,NEURIPS2025_38b445e9,Ma_2026_CVPR,Kim_2026_CVPR, xu2026rethinking}.
More recent approaches additionally incorporate textual guidance into pre-LLM selection~\cite{zhang2025cdpruner,liu2026hiprune, yu2026visiontrim,Du_2026_CVPR}.
ST-Pruner belongs to this text-guided pre-LLM setting, but differs in how its structural and semantic selection signals are constructed, as discussed next.

\subsection{Structural and Semantic Signals for Token \mbox{Selection}}

Two forms of evidence are particularly relevant to pre-LLM token selection: preservation of visual structure and query relevance.
For structural selection, prior methods rely on visual attention, feature redundancy, diversity, spatial coverage, approximation error, or token sensitivity~\cite{Yang_2025_CVPR,Zhang_2025_ICCV, liu2026hiprune,Alvar_2025_CVPR,NEURIPS2025_38b445e9, Ma_2026_CVPR,Kim_2026_CVPR}.
These criteria primarily derive importance from attention distributions or relations among visual tokens.
In contrast, ST-Pruner uses an attention-independent spectral criterion, deriving token-wise saliency from each patch's contribution to the dominant singular components of the visual feature matrix.

Query relevance provides a complementary source of evidence.
TRIM~\cite{song2025less} exploits pretrained vision--language similarity, while recent pre-LLM methods incorporate textual guidance through instruction relevance, textual similarity, or text-guided complementation~\cite{zhang2025cdpruner,yu2026visiontrim}.
Prior CLIP interpretability studies further report that local patch--text responses can exhibit counter-intuitive spatial behavior due to pooling-induced semantic shift~\cite{li2022exploring}.
Recent work has also explored reversed CLIP similarity for grounding-oriented visual token pruning~\cite{lee2026clip}.
ST-Pruner instead uses inverted patch--text similarity as a pre-LLM surrogate for query-conditioned token selection and combines it with SVD-derived spectral saliency.

\section{Method}

\label{sec:method}

\begin{figure*}[t]
    \centering
    \includegraphics[
        trim={10pt 5pt 10pt 15pt},
        width=\textwidth]{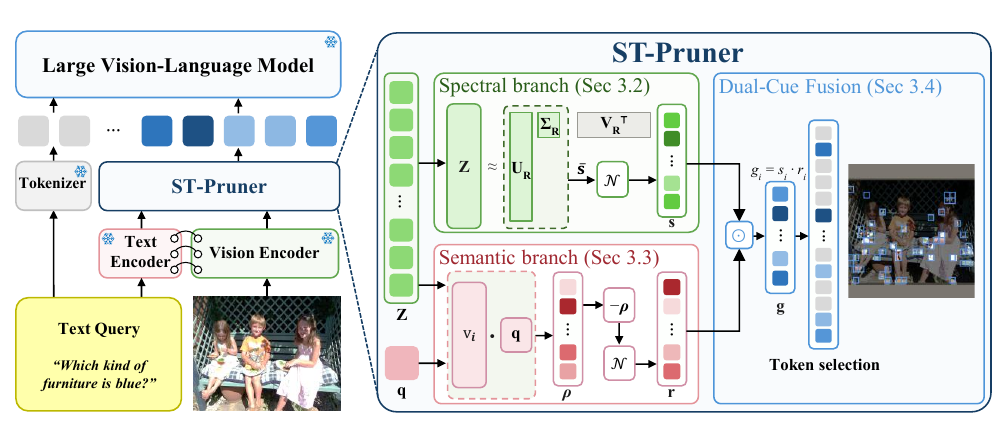}
    \caption{
        \textbf{Overview of ST-Pruner.}
        Given pre-LLM visual features $\mathbf{Z}$, the spectral branch derives its saliency score from a randomized low-rank SVD, while the semantic branch computes inverted patch--text relevance using a query embedding from the corresponding frozen text encoder of the paired vision--language model.
        Their parameter-free multiplicative fusion determines the top-$K$ visual tokens, which are retained in their original sequence order before LLM inference.
        The textual query is independently tokenized for the standard language-model input.
    }
    \label{fig:pipeline}
\end{figure*}

We present \textbf{ST-Pruner}, a training-free framework that performs visual token selection entirely before LLM inference.
Figure~\ref{fig:pipeline} illustrates the overall pipeline, followed by the spectral, semantic, and fusion components described below.

\subsection{Problem Formulation and Overview}
\label{sec:overview}

For notational simplicity, we first consider a single visual group, such as an image, a high-resolution view, or a video frame.
Given a visual input group $\mathbf{X}$, a pretrained vision encoder $f_v$ produces $N$ spatial patch features,
\begin{equation}
\mathbf{Z}=f_v(\mathbf{X}) =[\mathbf{z}_1,\ldots,\mathbf{z}_N]^\top \in\mathbb{R}^{N\times d}, \label{eq:visual_features} 
\end{equation}
where the class token, when present, is excluded from $\mathbf{Z}$.
For multi-view images or videos, scoring, normalization, and token selection are performed independently within each view or frame, and the retained groups are concatenated afterward.

Let $T$ denote the textual query.
In a standard LVLM, the visual features are transformed through the model-specific visual interface, typically a multimodal projector, and provided to the LLM together with the tokenized representation of $T$.
When a paired vision--language encoder is available, $T$ is also mapped to its pretrained shared embedding space for pre-LLM token scoring.

ST-Pruner selects a subset of visual indices within each group,
\begin{equation}
\mathcal{S}\subseteq\{1,\ldots,N\}, \qquad |\mathcal{S}|=K<N,
\end{equation}
where $K$ denotes the per-group token budget in the formulation below.
For multi-group inputs, the reported total budget is the sum of the retained tokens across all groups.
For projector-based LLaVA-family models, the pre-projector ViT features $\mathbf{Z}$ are used for token scoring.
The selected indices are applied to the corresponding model-specific visual representations when constructing the final LLM input.

\subsection{Spectral Saliency}
\label{sec:structure}

As discussed in Section~\ref{sec:intro}, \texttt{[CLS]}-based selection can become scattered under aggressive compression.
We therefore avoid relying on a single global attention distribution and instead characterize each patch by its participation in the dominant low-rank structure of $\mathbf{Z}$.
We compute a randomized rank-$R$ SVD approximation~\cite{halko2011finding}: \begin{equation} \mathbf{Z} \approx \mathbf{U}_R \boldsymbol{\Sigma}_R \mathbf{V}_R^\top, \label{eq:lowrank_svd}  \end{equation}
where $R<\min(N,d)$ denotes the target spectral rank, $\mathbf{U}_R\in\mathbb{R}^{N\times R}$, and $\mathbf{V}_R\in\mathbb{R}^{d\times R}$.
The diagonal matrix $\boldsymbol{\Sigma}_R=\mathrm{diag}(\sigma_1,\ldots,\sigma_R)$ contains the approximate leading singular values.
Here, $U_{ik}$ denotes the coefficient associated with patch $i$ in the $k$-th left singular vector, the columns of $\mathbf{V}_R$ represent the corresponding directions in feature space, and $\sigma_k$ measures the strength of each component.

We define the raw spectral saliency score of patch $i$ as 
\begin{equation} \bar{s}_i = \sum_{k=1}^{R} \sqrt{\sigma_k}\,U_{ik}^{2}. 
\label{eq:svd_score} 
\end{equation}

The squared coefficient $U_{ik}^{2}$ measures the participation of patch $i$ in the $k$-th dominant component while making the score invariant to the arbitrary sign of the corresponding singular vector.
We use $\sqrt{\sigma_k}$ weighting to preserve the ordering of component
importance while tempering the dominance of leading components relative
to direct $\sigma_k$ weighting.
For a token-score vector $\mathbf{a}=[a_1,\ldots,a_N]^\top$ within one visual group, we define min--max normalization as
\begin{equation}
[\mathcal{N}(\mathbf{a})]_i = \frac{a_i-\min_j a_j} {\max_j a_j-\min_j a_j+\epsilon}, \label{eq:minmax}
\end{equation}
where $\epsilon$ is a small constant for numerical stability. 
Let $\bar{\mathbf{s}}=[\bar{s}_1,\ldots,\bar{s}_N]^\top$.
The normalized spectral saliency is
\begin{equation}
s_i=[\mathcal{N}(\bar{\mathbf{s}})]_i, \qquad s_i\in[0,1]. \label{eq:spectral_score}
\end{equation}
Model-specific rank settings are provided in Section~\ref{sec:exp}.


\subsection{Pre-LLM Semantic Relevance}
\label{sec:relevance}

Spectral saliency is query-independent and therefore does not explicitly capture query relevance.
For models equipped with a paired vision--language encoder, we use its pretrained shared embedding space to obtain a query-conditioned signal before LLM inference.

Let $\phi(\cdot)$ denote the frozen visual projection paired with the vision encoder $f_v$, and let $\psi(\cdot)$ denote the corresponding frozen text encoder and text projection.
The visual projection maps each pre-projector patch feature $\mathbf{z}_i$ to the shared vision--language embedding $\mathbf{v}_i=\phi(\mathbf{z}_i)$.
The corresponding text encoder produces an EOS-token representation, which is mapped by the text projection to $\mathbf{q}=\psi(T)$.
Both $\mathbf{v}_i$ and $\mathbf{q}$ are $\ell_2$-normalized.
We then compute patch--text cosine similarity as
\begin{equation}
\rho_i = \mathbf{v}_i^\top \mathbf{q}.
\label{eq:sim}
\end{equation}

The projection $\phi$ is part of the paired pretrained vision--language model and is distinct from the LVLM multimodal projector.
It is used only for relevance scoring, so no LLM representation or decoder-side attention is required.

Such pretrained vision--language encoders are optimized primarily with global alignment objectives rather than explicitly calibrating local patch--text responses for token importance.
For CLIP, prior interpretability work further reports pooling-induced semantic shift in such local responses~\cite{li2022exploring}.
These observations, together with the layer-wise evidence in Figure~\ref{fig:fig2}(a,c), motivate our use of inverted local similarity as an empirical relevance surrogate.
Let $\boldsymbol{\rho}=[\rho_1,\ldots,\rho_N]^\top$. 
We define
\begin{equation}
r_i = [\mathcal{N}(-\boldsymbol{\rho})]_i, \qquad r_i\in[0,1].
\label{eq:relevance}
\end{equation}
A larger $r_i$ therefore indicates higher token importance under the inverted similarity surrogate.
For architectures that do not expose a paired pre-LLM text representation, the semantic branch is omitted and ST-Pruner operates using spectral saliency alone.
We evaluate this spectral-only setting on Qwen2.5-VL in Section~\ref{sec:exp}.


\subsection{Dual-Cue Fusion and Token Selection}
\label{sec:fusion}

The spectral score $s_i$ reflects a patch's participation in dominant visual structure, whereas the semantic score $r_i$ encodes query-conditioned relevance under the inverted similarity surrogate.

We combine the two normalized scores using a parameter-free multiplicative rule:
\begin{equation}
g_i = s_i \cdot r_i.
\label{eq:fusion}
\end{equation}
Collecting the token-wise fusion scores, let $\mathbf{g}=[g_1,\ldots,g_N]^\top$.
Multiplicative fusion favors agreement between the two cues: a patch receives a high final score only when it is supported by both structural and semantic evidence.
Unlike weighted additive fusion, this formulation introduces no additional balancing coefficient between the two normalized scores.

The retained token set is then obtained by
\begin{equation}
\mathcal{S}
=
\operatorname{TopK}(\mathbf{g},K),
\qquad
|\mathcal{S}|=K.
\label{eq:selection}
\end{equation}
Here, $\operatorname{TopK}(\mathbf{g},K)$ returns the indices of the
$K$ largest entries of $\mathbf{g}$.

The selected indices are sorted by their original token positions before constructing the retained visual-token sequence, so pruning does not reorder the surviving tokens.
ST-Pruner requires neither additional training nor learnable fusion parameters, and no LLM representation is required for token scoring.


\subsection{Computational Complexity}
\label{sec:complexity}

For $\mathbf{Z}\in\mathbb{R}^{N\times d}$ and $R\ll\min(N,d)$, the randomized rank-$R$ SVD has an approximate computational cost of $O(NdR)$ for a fixed number of subspace iterations, followed by $O(NR)$ computation for the spectral score.
Thus, the spectral branch has leading complexity
\begin{equation}
\mathcal{C}_{\mathrm{spec}}
=
O(NdR + NR).
\label{eq:spectral_complexity}
\end{equation}

The semantic branch requires a single frozen text-encoder forward pass, followed by visual projection and patch--text similarity computation, while the final fusion consists only of an element-wise product between the two token-wise scores.
Since both scoring branches are completed before decoder execution, each visual group is reduced from $N$ to $K$ tokens before the retained groups are concatenated into the LLM input.
Because self-attention scales quadratically with sequence length, reducing the visual prefix lowers the attention cost throughout all LLM layers~\cite{NIPS2017_3f5ee243}.
End-to-end savings including scoring overhead are evaluated empirically in Section~\ref{sec:exp}.

\section{Experiments}
\label{sec:exp}

\setcounter{dbltopnumber}{1}


\begin{table*}[!t]
\centering
\caption{\textbf{Performance comparison on LLaVA-1.5-7B.}
Avg. denotes mean performance retention relative to the displayed
full-token upper bound over the available metrics.}
\label{tab:llava15}
\footnotesize
\renewcommand{\arraystretch}{0.91}
\setlength{\tabcolsep}{10.6pt}
\begin{tabular}{l|cccccccc|c}
\toprule
Method & VQA$^{\mathrm{V2}}$ & GQA & VizWiz & SQA &
VQA$^{\mathrm{Text}}$ & POPE & MME$^{\mathrm{P}}$ & MMB & Avg. \\
\midrule
\rowcolor{gray!15}
\multicolumn{10}{c}{\emph{Upper Bound, 576 Tokens} (\textbf{100\%})} \\
LLaVA-1.5-7B & 78.5 & 61.9 & 50.1 & 69.5 & 58.2 & 85.9 & 1506.5 & 64.7 & 100.0 \\
\rowcolor{gray!15}
\multicolumn{10}{c}{\emph{Retain 192 Tokens} (\(\downarrow 66.7\%\))} \\
FastV \venue{ECCV'24}       & 67.1 & 52.7 & 50.8 & 67.3 & 52.5 & 64.8 & --     & 61.2 & 89.9 \\
SparseVLM \venue{ICML'25}   & 75.6 & 57.6 & 50.5 & 69.1 & 56.1 & 83.6 & --     & 62.5 & 97.1 \\
VisionZip \venue{CVPR'25}   & 76.8 & 59.3 & 50.5 & 68.9 & 57.3 & 85.3 & 1432.7 & 63.0 & 98.0 \\
DivPrune \venue{CVPR'25}    & 75.5 & 60.0 & \textbf{52.4} & \textbf{69.6} & 56.6 & 87.2 & 1427.5 & 61.5 & 98.3 \\
VisPruner \venue{ICCV'25}   & 77.0 & 59.6 & 51.9 & 68.5 & \textbf{57.7} & 85.9 & \textbf{1457.7} & 63.1 & 98.7 \\
ApET \venue{CVPR'26}        & 76.2 & 60.2 & 50.2 & 68.5 & 54.4 & 86.3 & 1439.0 & \textbf{63.4} & 97.6 \\
\rowcolor{oursbg}
\textbf{ST-Pruner (Ours)}   & \textbf{77.2} & \textbf{60.3} & 52.3 & 69.4 & 57.6 & \textbf{87.3} & 1449.9 & 62.5 & \textbf{99.2} \\
\rowcolor{gray!15}
\multicolumn{10}{c}{\emph{Retain 128 Tokens} (\(\downarrow 77.8\%\))} \\
FastV \venue{ECCV'24}       & 61.8 & 49.6 & 51.3 & 60.2 & 50.6 & 59.6 & 1368.9 & 56.1 & 85.2 \\
SparseVLM \venue{ICML'25}   & 73.8 & 56.0 & 51.4 & 67.1 & 54.9 & 80.5 & 1399.3 & 60.0 & 94.7 \\
VisionZip \venue{CVPR'25}   & 75.6 & 57.6 & 50.0 & 68.9 & 56.8 & 83.2 & 1436.3 & 62.0 & 96.7 \\
DivPrune \venue{CVPR'25}    & 76.0 & \textbf{59.4} & \textbf{52.8} & 68.6 & 55.9 & 87.0 & 1405.1 & 61.5 & 97.8 \\
VisPruner \venue{ICCV'25}   & 75.8 & 58.4 & 52.2 & 68.2 & \textbf{57.1} & 83.8 & 1436.7 & 62.2 & 97.5 \\
ApET \venue{CVPR'26}        & 75.1 & 58.9 & 50.7 & 68.7 & 53.9 & 86.1 & \textbf{1461.9} & \textbf{62.3} & 97.1 \\
\rowcolor{oursbg}
\textbf{ST-Pruner (Ours)}   & \textbf{76.2} & \textbf{59.4} & 52.7 & \textbf{69.4} & \textbf{57.1} & \textbf{87.3} & 1432.3 & 62.2 & \textbf{98.6} \\
\rowcolor{gray!15}
\multicolumn{10}{c}{\emph{Retain 64 Tokens} (\(\downarrow 88.9\%\))} \\
FastV \venue{ECCV'24}       & 55.0 & 46.1 & 50.8 & 51.1 & 47.8 & 48.0 & 973.5  & 48.0 & 74.5 \\
SparseVLM \venue{ICML'25}   & 68.2 & 52.7 & 50.1 & 62.2 & 51.8 & 75.1 & 1190.4 & 56.2 & 88.0 \\
VisionZip \venue{CVPR'25}   & 72.4 & 55.1 & 51.9 & 69.0 & 55.5 & 77.0 & 1380.5 & 60.1 & 94.2 \\
DivPrune \venue{CVPR'25}    & \textbf{74.1} & \textbf{57.5} & \textbf{53.6} & 68.0 & 54.5 & 85.5 & 1334.7 & 60.1 & 95.8 \\
VisPruner \venue{ICCV'25}   & 72.7 & 55.6 & 53.1 & 68.4 & \textbf{55.6} & 78.1 & 1357.3 & 60.0 & 94.5 \\
ApET \venue{CVPR'26}        & 72.5 & 56.9 & 51.9 & 68.9 & 53.0 & 84.4 & \textbf{1473.3} & \textbf{61.2} & 96.1 \\
\rowcolor{oursbg}
\textbf{ST-Pruner (Ours)}   & 73.3 & 56.8 & 52.1 & \textbf{69.8} & 55.5 & \textbf{86.5} & 1384.4 & 59.8 & \textbf{96.3} \\
\bottomrule
\end{tabular}
\end{table*}

\begin{table}[t]
\centering
\caption{\textbf{Performance comparison on LLaVA-NeXT-7B.}
Avg. denotes mean retention relative to the reproduced full-token upper
bound over the displayed metrics.
LLaVA-NeXT uses a variable number of views, and the displayed token
counts denote maxima for five-view inputs.}
\label{tab:llavanext}
\scriptsize
\renewcommand{\arraystretch}{0.91}
\setlength{\tabcolsep}{1.6pt}

\begin{tabular}{l|ccccccc|c}
\toprule
Method &
VQA$^{\mathrm{V2}}$ & GQA & VizWiz & SQA &
VQA\textsuperscript{\tiny Text} & POPE & MME$^{\mathrm{P}}$ & Avg. \\
\midrule

\rowcolor{gray!15}
\multicolumn{9}{c}{\emph{Upper Bound, 2880 Tokens} (\textbf{100\%})} \\
LLaVA-NeXT-7B
& 81.3 & 62.7 & 54.7 & 72.1 & 60.3 & 86.5 & 1501.1 & 100.0 \\

\rowcolor{gray!15}
\multicolumn{9}{c}{\emph{Retain 640 Tokens} (\(\downarrow 77.8\%\))} \\
FastV
& 77.0 & 58.9 & 53.9 & 67.4 & 58.1 & 79.5 & 1412.6 & 94.7 \\
SparseVLM
& 79.2 & 61.2 & 53.6 & 67.6 & 59.7 & 85.3 & 1456.8 & 97.3 \\
VisionZip
& 79.1 & 61.2 & 57.1 & 68.1 & 59.9 & 86.0 & 1493.4 & 98.9 \\
DivPrune
& 78.9 & 61.5 & 54.9 & \textbf{71.8} & 55.5 & 86.0 & 1466.9 & 97.8 \\
VisPruner
& \textbf{79.8} & 61.5 & \textbf{57.3} & 67.9 & 59.4 & 85.9 & 1470.3 & 98.7 \\
ApET
& 79.2 & \textbf{63.6} & 55.0 & \textbf{71.8} & 56.6 & 86.8 & \textbf{1513.9} & 99.1 \\
\rowcolor{oursbg}
\textbf{ST-Pruner (Ours)}
& 79.1 & 62.0 & 55.8 & \textbf{71.8} & \textbf{60.0} & \textbf{87.0} & 1486.2 & \textbf{99.6} \\

\rowcolor{gray!15}
\multicolumn{9}{c}{\emph{Retain 320 Tokens} (\(\downarrow 88.9\%\))} \\
FastV
& 61.5 & 49.8 & 51.3 & 66.6 & 52.2 & 49.5 & 1099.0 & 79.7 \\
SparseVLM
& 74.6 & 57.9 & 54.2 & 67.2 & 56.5 & 76.9 & 1386.1 & 93.0 \\
VisionZip
& 76.2 & 58.9 & 56.2 & 67.5 & \textbf{58.8} & 82.3 & 1397.1 & 95.7 \\
DivPrune
& \textbf{76.6} & 59.8 & 53.7 & \textbf{71.8} & 53.6 & 83.4 & 1415.1 & 95.3 \\
VisPruner
& 75.7 & 58.9 & \textbf{57.3} & 68.1 & \textbf{58.8} & 81.3 & 1424.6 & 96.1 \\
ApET
& 75.8 & \textbf{61.3} & 53.2 & 71.7 & 54.2 & 85.1 & 1435.7 & 95.9 \\
\rowcolor{oursbg}
\textbf{ST-Pruner (Ours)}
& 76.3 & 60.4 & 54.6 & 71.5 & 58.6 & \textbf{86.1} & \textbf{1448.2} & \textbf{97.5} \\

\rowcolor{gray!15}
\multicolumn{9}{c}{\emph{Retain 160 Tokens} (\(\downarrow 94.4\%\))} \\
VisionZip
& 71.4 & 55.2 & 55.5 & 67.9 & 55.0 & 74.9 & 1327.8 & 91.1 \\
DivPrune
& \textbf{73.9} & 57.9 & 53.6 & \textbf{72.1} & 52.6 & 79.6 & 1358.5 & 93.0 \\
VisPruner
& 70.6 & 56.1 & \textbf{57.2} & 67.6 & \textbf{55.9} & 75.4 & 1321.5 & 91.8 \\
ApET
& 72.7 & \textbf{58.1} & 51.0 & 71.5 & 51.2 & 82.1 & \textbf{1361.3} & 92.1 \\
\rowcolor{oursbg}
\textbf{ST-Pruner (Ours)}
& 73.0 & \textbf{58.1} & 54.2 & 71.7 & \textbf{55.9} & \textbf{85.0} & 1329.4 & \textbf{94.4} \\

\bottomrule
\end{tabular}
\vspace{-3pt} 
\end{table}


\subsection{Experimental Setup}

\paragraph{Models and baselines.}
For image understanding, we evaluate ST-Pruner on the widely adopted
LLaVA family, using LLaVA-1.5-7B~\cite{liu2024llava15} as a standard
setting and LLaVA-NeXT-7B~\cite{liu2024llavanext} for substantially
longer multi-view visual inputs.
To assess cross-architecture transfer, we further evaluate the spectral
branch on Qwen2.5-VL-7B~\cite{qwen2025vl72b}.
For video understanding, we adopt
Video-LLaVA-7B~\cite{lin2024video} to evaluate ST-Pruner on
multi-frame visual inputs.
We compare against FastV~\cite{chen2024image},
SparseVLM~\cite{pmlr-v267-zhang25s},
VisionZip~\cite{Yang_2025_CVPR},
DivPrune~\cite{Alvar_2025_CVPR},
VisPruner~\cite{Zhang_2025_ICCV},
CDPruner~\cite{zhang2025cdpruner}, and
ApET~\cite{Ma_2026_CVPR}.
Published results are used when directly compatible with our setting.
Otherwise, we reproduce the official implementation.

\paragraph{Datasets.}
For image understanding, we evaluate on
VQAv2~\cite{goyal2017making},
GQA~\cite{hudson2019gqa},
VizWiz~\cite{gurari2018vizwiz},
ScienceQA-IMG~\cite{lu2022learn},
TextVQA~\cite{singh2019towards},
POPE~\cite{li2023evaluating},
MME~\cite{fu2026mme}, and
MMBench (MMB)~\cite{liu2024mmbench}.
For video understanding, we evaluate on
TGIF-QA~\cite{jang2017tgif},
MSVD-QA~\cite{xu2017video}, and
MSRVTT-QA~\cite{xu2017video}.
We report the standard benchmark metrics: accuracy for VQAv2, VizWiz, ScienceQA-IMG, TextVQA, and MMBench, exact match for GQA, F1 for POPE, and MME Perception (MME$^{\mathrm{P}}$).
For video QA, we report GPT-3.5-turbo-judged accuracy following the standard Video-LLaVA protocol.
For aggregate comparison, each benchmark score is normalized by its corresponding full-token reference, and we report the mean retention over the displayed metrics.

\paragraph{Implementation details.}
LLaVA-1.5 retains 192/128/64 of 576 visual tokens, LLaVA-NeXT retains 128/64/32 tokens per 576-token view, corresponding to up to 640/320/160 tokens over five views, and Video-LLaVA retains 17 tokens per 256-token frame, yielding 136 tokens over eight frames.
Unless otherwise specified, ST-Pruner uses ViT-space features, randomized low-rank SVD, direct multiplicative fusion, and $R=N/8$ within each visual group.
All model parameters remain frozen.
Further model-specific implementation details are provided in the supplementary material.



\subsection{Main Results on Image LVLMs}

\paragraph{LLaVA-1.5.}
We first evaluate ST-Pruner on LLaVA-1.5-7B, a widely used setting for visual token pruning.
Following prior work, we consider three token budgets of 192, 128, and 64 out of the original 576 visual tokens.
As shown in Table~\ref{tab:llava15}, ST-Pruner achieves the highest average performance retention at all three compression levels, preserving $99.2\%$, $98.6\%$, and $96.3\%$ of the full-token reference, respectively.
Reducing the token budget threefold from 192 to 64 decreases average retention by only $2.9$ percentage points, while removing $88.9\%$ of the original visual tokens.


\paragraph{LLaVA-NeXT.}
We next examine whether ST-Pruner remains effective for substantially longer visual sequences.
Unlike LLaVA-1.5, LLaVA-NeXT processes up to five 576-token views, yielding up to 2880 visual tokens.
Table~\ref{tab:llavanext} shows that ST-Pruner retains $99.6\%$, $97.5\%$, and $94.4\%$ of full-token performance with only 640, 320, and 160 tokens, respectively.
These results show that pre-LLM selection remains effective for long multi-view visual sequences.



\subsection{Generalization Across Modalities and \mbox{Architectures}}

\paragraph{Video-LLaVA.}

We extend ST-Pruner to multi-frame video inputs using Video-LLaVA-7B, which represents eight frames with 256 visual tokens per frame.
At approximately $93.4\%$ visual-token reduction, ST-Pruner retains $95.4\%$ of the full-token reference performance.
The spectral branch alone retains $93.7\%$, while semantic relevance raises average retention to $95.4\%$, exceeding VisionZip by $4.4$ percentage points under the same aggressive reduction setting.


\begin{table}[!t]
\centering
\caption{\textbf{Performance comparison on Video-LLaVA-7B.}
Avg. denotes mean performance retention over the three video benchmarks
relative to the displayed full-token reference.}
\label{tab:video}
\footnotesize
\renewcommand{\arraystretch}{0.97}
\setlength{\tabcolsep}{5.5pt}
\begin{tabular}{l|cccc}
\toprule
Method & TGIF & MSVD & MSRVTT & Avg. \\
\midrule
\rowcolor{gray!15}
\multicolumn{5}{c}{\emph{Upper Bound, 2048 Tokens} (\textbf{100\%})} \\
Video-LLaVA-7B & 47.1 & 69.8 & 56.7 & 100.0 \\
\rowcolor{gray!15}
\multicolumn{5}{c}{\emph{Retain 136 Tokens}
(\(\downarrow 93.4\%\))} \\
FastV \venue{ECCV'24}       & 23.1 & 38.0 & 19.3 & 45.8 \\
SparseVLM \venue{ICML'25}   & 44.7 & \textbf{68.2} & 31.0 & 82.4 \\
VisionZip \venue{CVPR'25}   & 42.4 & 63.5 & 52.1 & 91.0 \\
\rowcolor{spectralbg}
\textbf{ST-Pruner (Spectral)}
& 43.6 & 66.9 & 52.5 & 93.7 \\
\rowcolor{oursbg}
\textbf{ST-Pruner (Ours)}
& \textbf{44.9} & 66.3 & \textbf{54.4} & \textbf{95.4} \\
\bottomrule
\end{tabular}
\end{table}

\begin{table}[!t]
\centering
\caption{\textbf{Cross-architecture comparison on Qwen2.5-VL-7B.}
ST-Pruner (Spectral) denotes the proposed spectral criterion without
the semantic branch.}
\label{tab:qwen}
\footnotesize
\renewcommand{\arraystretch}{0.94}
\setlength{\tabcolsep}{6.0pt}
\begin{tabular}{l|cccc}
\toprule
Method & GQA & SQA & POPE & Avg. \\
\midrule
\rowcolor{gray!15}
\multicolumn{5}{c}{\emph{Upper Bound, 1296 Tokens} (\textbf{100\%})} \\
Qwen2.5-VL-7B & 62.0 & 87.8 & 88.2 & 100.0 \\

\rowcolor{gray!15}
\multicolumn{5}{c}{\emph{Retain 512 Tokens} (\(\downarrow 60.5\%\))} \\
VisionZip \venue{CVPR'25} & 58.1 & 86.2 & 83.3 & 95.4 \\
DivPrune \venue{CVPR'25}  & 61.3 & \textbf{87.9} & 87.4 & 99.4 \\
VisPruner \venue{ICCV'25} & 60.8 & 87.3 & 86.7 & 98.6 \\
CDPruner \venue{NeurIPS'25} & 60.3 & 86.0 & 87.1 & 98.0 \\
ApET \venue{CVPR'26}      & \textbf{62.0} & 86.8 & 87.8 & 99.5 \\
\rowcolor{spectralbg}
\textbf{ST-Pruner (Spectral)}
& 61.8 & 86.8 & \textbf{88.4} & \textbf{99.6} \\

\rowcolor{gray!15}
\multicolumn{5}{c}{\emph{Retain 256 Tokens} (\(\downarrow 80.2\%\))} \\
VisionZip \venue{CVPR'25} & 55.6 & 85.5 & 79.7 & 92.5 \\
DivPrune \venue{CVPR'25}  & 60.5 & \textbf{87.2} & 86.8 & \textbf{98.4} \\
VisPruner \venue{ICCV'25} & 59.5 & 86.9 & 85.5 & 97.3 \\
CDPruner \venue{NeurIPS'25} & 58.8 & 85.2 & 86.0 & 96.5 \\
ApET \venue{CVPR'26}      & 60.5 & 86.2 & \textbf{87.1} & 98.2 \\
\rowcolor{spectralbg}
\textbf{ST-Pruner (Spectral)}
& \textbf{60.8} & 85.9 & \textbf{87.1} & 98.2 \\

\rowcolor{gray!15}
\multicolumn{5}{c}{\emph{Retain 128 Tokens} (\(\downarrow 90.1\%\))} \\
VisionZip \venue{CVPR'25} & 52.1 & 84.1 & 75.6 & 88.5 \\
DivPrune \venue{CVPR'25}  & \textbf{58.9} & \textbf{86.4} & \textbf{85.3} & \textbf{96.7} \\
VisPruner \venue{ICCV'25} & 57.5 & 85.7 & 82.5 & 94.6 \\
CDPruner \venue{NeurIPS'25} & 56.4 & 84.2 & 83.2 & 93.7 \\
ApET \venue{CVPR'26}      & 57.6 & 85.6 & 84.0 & 95.2 \\
\rowcolor{spectralbg}
\textbf{ST-Pruner (Spectral)}
& 57.9 & 85.3 & 84.4 & 95.4 \\
\bottomrule
\end{tabular}
\vspace{-4pt} 
\end{table}


\paragraph{Qwen2.5-VL.}
\label{sec:qwen}

Qwen2.5-VL-7B does not expose a CLIP-style text encoder paired with its vision encoder for pre-LLM patch--text scoring, so we evaluate the spectral branch in isolation.
Using $1008\times1008$ inputs, all entries in Table~\ref{tab:qwen} are reproduced under the same setting.
ST-Pruner (Spectral) achieves the highest average retention at 512 tokens ($99.6\%$), remains competitive at 256 tokens ($98.2\%$), and retains $95.4\%$ at 128 tokens.
These results show that the proposed spectral criterion transfers beyond the LLaVA-family architectures used by the full ST-Pruner.

\begin{table}[t]
\centering
\caption{\textbf{Signal ablation on LLaVA-1.5-7B.}
$+\mathrm{sim}$ and $-\mathrm{sim}$ denote conventional and inverted
patch--text similarity, respectively.
Spectral-only denotes the spectral branch of ST-Pruner.}
\label{tab:signal_ablation}
\footnotesize
\renewcommand{\arraystretch}{1.1}
\setlength{\tabcolsep}{5.5pt}
\begin{tabular}{l|cccc}
\toprule
Variant & GQA & POPE & MME$^{\mathrm{P}}$ & VQA$^{\mathrm{Text}}$ \\
\midrule
\rowcolor{gray!15}
\multicolumn{5}{c}{\emph{Retain 128 Tokens}} \\
\rowcolor{semanticbg}
Semantic-only ($+\mathrm{sim}$) & 39.5 & 1.5 & 626.0 & 42.7 \\
\rowcolor{semanticbg}
Semantic-only ($-\mathrm{sim}$) & \textbf{59.6} & 86.9 & 1407.0 & 53.9 \\
\rowcolor{spectralbg}
Spectral-only & 58.5 & 85.9 & 1377.2 & 57.0 \\
\rowcolor{oursbg}
\textbf{ST-Pruner (Ours)} & 59.4 & \textbf{87.3} & \textbf{1432.3} & \textbf{57.1} \\

\rowcolor{gray!15}
\multicolumn{5}{c}{\emph{Retain 64 Tokens}} \\
\rowcolor{semanticbg}
Semantic-only ($+\mathrm{sim}$) & 38.9 & 1.7 & 633.4 & 42.2 \\
\rowcolor{semanticbg}
Semantic-only ($-\mathrm{sim}$) & \textbf{57.9} & 86.4 & 1346.2 & 51.4 \\
\rowcolor{spectralbg}
Spectral-only & 54.3 & 82.1 & 1269.9 & 55.3 \\
\rowcolor{oursbg}
\textbf{ST-Pruner (Ours)} & 56.8 & \textbf{86.5} & \textbf{1384.4} & \textbf{55.5} \\
\bottomrule
\end{tabular}
\end{table}

\begin{table}[!t]
\centering
\vspace{-3pt}
\caption{\textbf{Robustness to fusion strategy and spectral rank.}
Values denote mean performance retention over GQA, POPE, TextVQA,
MME$^{\mathrm{P}}$, and MMB, normalized by the matched full-token
reproduction used for these ablations.
For entropy-weighted fusion, we report the strongest setting
($\tau=0.5$) from the full temperature sweep.
Complete per-benchmark results are provided in the supplementary
material.}
\label{tab:design_robustness}
\footnotesize
\renewcommand{\arraystretch}{0.96}
\setlength{\tabcolsep}{10pt}
\begin{tabular}{l|cc}
\toprule
Setting & $K=128$ & $K=64$ \\
\midrule
\rowcolor{gray!15}
\multicolumn{3}{c}{\emph{Fusion Strategy}} \\
Entropy-weighted ($\tau=0.5$) & 97.2 & 94.4 \\
\rowcolor{oursbg}
Direct product (Ours) & \textbf{97.4} & \textbf{94.5} \\
\midrule
\rowcolor{gray!15}
\multicolumn{3}{c}{\emph{Spectral Rank $R$ with Direct Product}} \\
$R=18$  & 96.2 & 93.6 \\
$R=36$  & 97.0 & 94.2 \\
\rowcolor{oursbg}
$\mathbf{R=72}$ \textbf{(Ours)} & \textbf{97.4} & \textbf{94.5} \\
$R=144$ & 97.0 & 94.0 \\
$R=288$ & 96.9 & 93.8 \\
\bottomrule
\end{tabular}
\vspace{-3pt} 
\end{table}

\subsection{Ablation Study}

\paragraph{Complementarity and semantic inversion.}
Table~\ref{tab:signal_ablation} examines whether inverted similarity is
preferable to conventional positive similarity and whether spectral and
semantic evidence are complementary.
Positive similarity collapses under aggressive pruning, whereas
inversion dramatically improves token selection. For example, POPE
rises from $1.5$ to $86.9$ at $K=128$.
This reversal supports treating inverted local similarity as an empirical
pre-LLM relevance surrogate rather than raw CLIP patch similarity as a
calibrated token-importance score.
The two cues are also complementary: semantic relevance is stronger on
GQA, whereas spectral saliency better preserves TextVQA performance,
and their fusion gives the best POPE, MME$^{\mathrm{P}}$, and TextVQA
results at both budgets.

\begin{table*}[t]
\centering
\caption{\textbf{Inference efficiency on LLaVA-NeXT-7B.}
Results are measured over the full POPE evaluation set of 8,910 samples
on a single NVIDIA V100 GPU under FP16.
The cross-method comparison uses a 320-token target budget.
Prefill and total latency include method-specific pruning and scoring
overhead.
Total denotes end-to-end generation latency with a maximum generation
length of 16 tokens.
Values in parentheses denote speedup or FLOPs reduction factors relative to the unpruned model.}
\label{tab:efficiency}

\scriptsize
\renewcommand{\arraystretch}{1.28}
\setlength{\tabcolsep}{12.2pt}

\begin{tabular}{l|c|c|c|c|c|c|c}
\toprule
Method &
Retain &
FLOPs (T) &
Prefill (ms) &
Decode (ms) &
Total (ms) &
KV (MB) &
POPE F1 \\
\midrule

\rowcolor{gray!15}
LLaVA-NeXT-7B
& 2880
& 34.92
& 679.2
& 48.8
& 727.9
& 1542
& 86.5 \\

SparseVLM \venue{ICML'25}
& 320\rlap{$^{\dagger}$}
& 7.81 {\scriptsize($\times4.5$)}
& 299.3 {\scriptsize($\times2.3$)}
& 44.1 {\scriptsize($\times1.1$)}
& 343.4 {\scriptsize($\times2.1$)}
& 359
& 76.9 \\

VisionZip \venue{CVPR'25}
& 320
& 4.53 {\scriptsize($\times7.7$)}
& \textbf{142.5} {\scriptsize($\times4.8$)}
& 36.0 {\scriptsize($\times1.4$)}
& \textbf{178.4} {\scriptsize($\times4.1$)}
& 200
& 82.3 \\

DivPrune \venue{CVPR'25}
& 320
& 4.53 {\scriptsize($\times7.7$)}
& 187.5 {\scriptsize($\times3.6$)}
& 33.6 {\scriptsize($\times1.5$)}
& 221.1 {\scriptsize($\times3.3$)}
& 200
& 83.4 \\

VisPruner \venue{ICCV'25}
& 320
& 4.53 {\scriptsize($\times7.7$)}
& 163.0 {\scriptsize($\times4.2$)}
& 34.9 {\scriptsize($\times1.4$)}
& 197.9 {\scriptsize($\times3.7$)}
& 200
& 81.3 \\

CDPruner \venue{NeurIPS'25}
& 320
& 4.53 {\scriptsize($\times7.7$)}
& 177.7 {\scriptsize($\times3.8$)}
& 34.2 {\scriptsize($\times1.4$)}
& 211.9 {\scriptsize($\times3.4$)}
& 200
& 85.8 \\

\rowcolor{spectralbg}
\textbf{ST-Pruner (Spectral)}
& 320
& 4.53 {\scriptsize($\times7.7$)}
& 165.4 {\scriptsize($\times4.1$)}
& \textbf{31.8} {\scriptsize($\times1.5$)}
& 197.2 {\scriptsize($\times3.7$)}
& 200
& 83.1 \\

\rowcolor{oursbg}
\textbf{ST-Pruner (Ours)}
& 320
& 4.53 {\scriptsize($\times7.7$)}
& 188.0 {\scriptsize($\times3.6$)}
& 36.0 {\scriptsize($\times1.4$)}
& 224.0 {\scriptsize($\times3.2$)}
& 200
& \textbf{86.1} \\

\bottomrule
\end{tabular}

\vspace{2pt}
{\scriptsize
$^{\dagger}$SparseVLM FLOPs and KV use the actual layer-wise token
trajectory under progressive in-LLM pruning, not the final token count.}

\vspace{5pt}
\end{table*}

\paragraph{Robustness to design choices.}

Table~\ref{tab:design_robustness} evaluates sensitivity to the fusion strategy and spectral rank.
As an adaptive alternative, entropy-weighted fusion gives greater influence to the lower-entropy cue, with $\tau$ controlling the weighting sharpness.
Even its strongest setting, $\tau=0.5$, matches but does not improve over direct multiplication, indicating limited sensitivity to the fusion rule.
Across $R=18$ to $288$, average retention varies by only $1.2$ percentage points at $K=128$ and $0.9$ points at $K=64$, showing similarly low sensitivity to the spectral rank.
The adopted $R=72$ achieves the best average retention among the evaluated ranks.
Further rank sweeps and qualitative analysis of weaker singular components are provided in the supplementary material.


\subsection{Efficiency}

To evaluate practical inference efficiency, we compare ST-Pruner with representative token-pruning methods on LLaVA-NeXT-7B over the full POPE evaluation set of 8,910 samples.
All experiments are conducted on a single NVIDIA V100 GPU under FP16, and the cross-method comparison uses a 320-token target budget.
At this budget, ST-Pruner reduces analytical LLM prefill FLOPs from $34.9$T to $4.5$T and the visual-prefix KV cache from 1542 MB to 200 MB.

As shown in Table~\ref{tab:efficiency}, all reported latencies include method-specific pruning and scoring overhead.
For ST-Pruner, this includes CLIP text encoding, patch--text relevance computation, randomized low-rank SVD, multiplicative fusion, and top-$K$ selection.
All components are executed within the timed generation call without precomputation.
For SparseVLM, which progressively removes tokens inside the LLM, FLOPs and KV cache are computed from the actual layer-wise token trajectory rather than from the final retained token count.

ST-Pruner achieves a $3.6\times$ prefill speedup and a $3.2\times$ end-to-end speedup while retaining a POPE F1 of $86.1$.
Its spectral-only variant reaches $165.4$ ms prefill and $197.2$ ms end-to-end latency, closely matching VisPruner at $163.0$ ms and $197.9$ ms while improving F1 from $81.3$ to $83.1$.
VisionZip is faster at $142.5$ ms prefill and $178.4$ ms end-to-end, but its F1 of $82.3$ is lower than that of the spectral-only variant.

The full ST-Pruner further improves F1 to $86.1$ while operating in a similar latency range to DivPrune and CDPruner.
DivPrune reaches $83.4$ F1 at $221.1$ ms total latency, while CDPruner reaches $85.8$ F1 at $211.9$ ms.
Relative to spectral-only selection, the semantic branch adds only
$22.6$ ms of prefill latency at $K=320$ while improving POPE F1 from
$83.1$ to $86.1$, corresponding to a $3.0$-point gain for an additional
$26.8$ ms of end-to-end latency.
Complete 640/320/160-token efficiency results are provided in the
supplementary material.

\FloatBarrier
\section{Conclusion}


We presented ST-Pruner as a pre-LLM alternative to decoder-dependent
visual token selection.
Rather than treating pruning solely as an attention-ranking problem,
our results show that dominant spectral structure provides an
attention-independent cue for preserving coherent visual evidence,
while inverted patch--text similarity supplies complementary
query-conditioned relevance.
Across image, video, and multi-view LVLMs, these cues enable effective
visual-token reduction while preserving strong downstream performance,
with ST-Pruner achieving a $3.2\times$ end-to-end speedup on
LLaVA-NeXT.
More broadly, our findings suggest that the geometry of visual
representations itself can provide useful signals for efficient
multimodal inference beyond conventional attention-based importance.

\paragraph{Limitations and future work.}

The semantic branch of ST-Pruner assumes access to a text encoder paired with the vision encoder for pre-LLM patch--text scoring, which is not exposed by all LVLM architectures.
In such cases, ST-Pruner currently operates using spectral saliency alone, as demonstrated on Qwen2.5-VL.
Extending query-conditioned pre-LLM selection to architectures without such paired text encoders and examining whether similar spectral structure persists across different vision encoders and multimodal representations are promising directions.

\paragraph{Additional analysis.}
Beyond the main paper, the supplementary material provides detailed evaluation protocols and complete fusion and spectral-rank sweeps (Secs.~6--7), feature-space and singular-value weighting ablations with qualitative spectral analyses (Sec.~8), and full efficiency results across multiple token budgets (Sec.~9).
\section*{Acknowledgments}

This work was supported by the Institute of Information \&
Communications Technology Planning \& Evaluation (IITP)
under the ICT Challenge and Advanced Network of HRD
(ICAN) program funded by the Korea government (MSIT)
(IITP-2026-RS-2024-00436528, 40\%),
the IITP under the Virtual Convergence Support Program
to Nurture the Best Talents grant funded by the Korea
government (MSIT) (IITP-2026-RS-2023-00254529, 30\%),
and the IITP under the Leading Generative AI Human
Resources Development grant funded by the Korea
government (MSIT) (IITP-2026-RS-2026-25544647, 30\%).

{\small
\bibliographystyle{ieeenat_fullname}
\bibliography{main}
}

\clearpage
\onecolumn

\begin{center}
    {\Large\bfseries
    Don’t Ask the LLM: Visual Token Pruning with Spectral Structure and
    Semantic Relevance\par}
    \vspace{0.5em}
    {\large Supplementary Material\par}
    \vspace{1.0em}
\end{center}


\setcounter{section}{5}
\setcounter{table}{0}
\setcounter{figure}{0}
\renewcommand{\thetable}{S\arabic{table}}
\renewcommand{\thefigure}{S\arabic{figure}}

\noindent
The supplementary material is organized as follows:
\begin{itemize}
    \renewcommand{\labelitemi}{\textendash}
    \setlength{\itemsep}{0pt}
    \setlength{\parskip}{0pt}
    \setlength{\parsep}{0pt}

    \item Section~\ref{sec:supp_eval}: dataset descriptions, model configurations,
    token budgets, and implementation details.

    \item Section~\ref{sec:supp_hparam}: fine-grained sensitivity analyses of
    the fusion strategy and spectral rank.

    \item Section~\ref{sec:supp_design}: additional ablations on feature space
    and singular-value weighting, together with qualitative spectral analysis.

    \item Section~\ref{sec:supp_efficiency}: complete inference-efficiency
    results and computational cost analysis.
\end{itemize}


\section{Evaluation Details}
\label{sec:supp_eval}

\subsection{Datasets}
\label{sec:supp_datasets}

We evaluate ST-Pruner on eight image benchmarks and three video benchmarks used in standard LLaVA and Video-LLaVA evaluation pipelines.
Unless otherwise stated, we follow the task-specific inference and evaluation protocols used by the corresponding backbone.

\subsubsection{Image Benchmarks}

\paragraph{VQAv2~\cite{goyal2017making}.}
VQAv2 evaluates open-ended visual question answering while reducing
language-only answer priors through adversarially balanced image--question
pairs with contrasting answers.
The benchmark is built on MSCOCO images and provides multiple human
reference answers per question.
We follow the standard LLaVA protocol and report VQA accuracy on the
test-dev split containing 107,394 image--question pairs.

\paragraph{GQA~\cite{hudson2019gqa}.}
GQA focuses on compositional visual reasoning over  objects, attributes, and relations derived from scene graphs.
We evaluate on the test-dev split containing 12,578 image--question pairs and report exact-match accuracy.

\paragraph{VizWiz~\cite{gurari2018vizwiz}.}
VizWiz contains real-world images and questions collected from blind users.
Its unconstrained images include challenging cases involving blur, poor
lighting, occlusion, and questions that may be unanswerable from the image.
We follow the standard LLaVA evaluation setting and report accuracy on the
8,000-example test-dev split.

\paragraph{ScienceQA-IMG~\cite{lu2022learn}.}
ScienceQA is a multimodal multiple-choice benchmark covering natural
science, social science, and language-related topics across diverse
topics, categories, and skills.
Following prior LLaVA evaluations, we use the image-associated subset of
the test set (SQA-IMG), containing 2,017 image--question pairs, and report
multiple-choice accuracy.

\paragraph{TextVQA~\cite{singh2019towards}.}
TextVQA measures question answering that requires recognizing and reasoning
over text embedded in natural images, including signs, advertisements, and
product packaging.
Answers may be read directly from visible text or require combining OCR
evidence with surrounding visual context.
We evaluate on the 5,000-example validation split and report the standard
TextVQA accuracy.

\paragraph{POPE~\cite{li2023evaluating}.}
POPE evaluates object hallucination through binary questions about whether a queried object is present in an image.
We report the average F1 score over the three standard sampling strategies.
The full evaluation set contains 8,910 image--question pairs.

\paragraph{MME~\cite{fu2026mme}.}
MME evaluates multimodal perception and cognition through a collection of
binary judgment tasks.
Following the main paper, we report the MME Perception score
(MME$^{\mathrm{P}}$), whose subtasks cover coarse- and fine-grained
recognition, counting, position, color, OCR, and related perceptual
capabilities.
The perception evaluation used here contains 2,374 image--question pairs.

\paragraph{MMBench~\cite{liu2024mmbench}.}
MMBench is a bilingual multiple-choice benchmark designed for fine-grained
evaluation of multimodal perception and reasoning.
Its capability taxonomy spans high-level perception and reasoning categories
and progressively finer visual-language abilities.
The benchmark further uses CircularEval, which cyclically permutes answer
choices to reduce option-position bias and tests whether the model answers a
question consistently across permutations.
We evaluate the English version (MMBench-EN), which contains 4,377
image--question pairs, using the standard LLaVA evaluation pipeline and
report circular accuracy.

\subsubsection{Video Benchmarks}

\paragraph{Video evaluation protocol.}
For TGIF-QA, MSVD-QA, and MSRVTT-QA, we follow the standard
Video-ChatGPT GPT-3.5-turbo evaluation protocol adopted by
Video-LLaVA~\cite{lin2024video}.
Generated answers are judged against the reference answers using the same correctness prompt and score-parsing procedure as the original evaluation pipeline.
Our evaluation implementation preserves this judging protocol while parallelizing API requests for efficient evaluation.
We evaluate the full evaluation set used by our implementation for each benchmark and report GPT-3.5-turbo-judged accuracy.

\paragraph{TGIF-QA~\cite{jang2017tgif}.}
TGIF-QA extends image-based visual question answering to animated GIFs and
requires both spatial and temporal understanding.
The benchmark contains approximately 72K GIFs and 165K crowdsourced
question--answer pairs across several task types.
We evaluate the FrameQA task.

\paragraph{MSVD-QA~\cite{xu2017video}.}
MSVD-QA is constructed from the Microsoft Research Video Description Corpus
and evaluates open-ended question answering over short video clips.
The benchmark is derived from 1,970 videos and contains approximately 50.5K
question--answer pairs.

\paragraph{MSRVTT-QA~\cite{xu2017video}.}
MSRVTT-QA is based on the larger MSR-VTT video-text corpus, which contains
10K videos with more diverse scenes and content than MSVD.
The derived QA benchmark contains approximately 243K question--answer pairs.
We use the standard Video-LLaVA evaluation protocol.

\subsection{Models and Implementation Details}
\label{sec:supp_models}

\paragraph{LLaVA-1.5-7B~\cite{liu2024llava15}.}
LLaVA-1.5 follows the standard LLaVA architecture, coupling a pretrained
CLIP ViT-L/14 vision encoder with a Vicuna-v1.5 language model through a
two-layer MLP vision--language projector.
Compared with the original LLaVA, LLaVA-1.5 increases the image resolution
to $336\times336$ and incorporates additional academic VQA data during
visual instruction tuning.
At this resolution, the CLIP vision tower produces a $24\times24$ grid of
576 spatial patch tokens before the multimodal projector.

\paragraph{LLaVA-NeXT-7B~\cite{liu2024llavanext}.}
LLaVA-NeXT, also referred to as LLaVA-1.6, extends the LLaVA-1.5 design
with dynamic high-resolution processing.
Its AnyRes strategy selects an image grid according to the input aspect
ratio, divides a high-resolution image into multiple $336\times336$ views,
and encodes the resulting visual groups with the same vision tower before
their representations are assembled for the LLM.
This design preserves substantially more local image detail than the
single-view LLaVA-1.5 input and is particularly relevant to OCR and
fine-grained visual reasoning.
Under the configuration used in our experiments, an image contains up to
five independently encoded 576-token visual groups, corresponding to at
most 2880 visual tokens.

\paragraph{Video-LLaVA-7B~\cite{lin2024video}.}
Video-LLaVA extends the LLaVA-style framework to joint image and video
understanding through the ``alignment before projection'' design.
It uses LanguageBind visual encoders whose image and video representations
are pretrained in a shared language-aligned feature space, followed by a
shared projection interface to the language model.
For video input, multiple sampled frames are encoded and their visual-token
representations are concatenated before LLM inference.
Following the standard Video-LLaVA evaluation setting used in our
experiments, each video is represented by eight frames with 256 visual
tokens per frame.

\paragraph{Qwen2.5-VL-7B~\cite{qwen2025vl72b}.}
Qwen2.5-VL uses a native dynamic-resolution Vision Transformer together
with the Qwen2.5 language model.
Its vision encoder incorporates window attention for efficient
high-resolution processing, while the multimodal architecture supports
variable-resolution images and videos through a model-specific visual
merging and projection interface.
Unlike the LLaVA-family settings above, Qwen2.5-VL does not expose a
CLIP-style text encoder paired with its vision encoder for pre-LLM
patch--text scoring.
We therefore use Qwen2.5-VL as a cross-architecture evaluation of the
spectral branch.
For controlled comparison, we use the fixed $1008\times1008$ setting from
the main paper, which produces 1296 visual tokens.

\paragraph{Token budgets and spectral settings.}
For LLaVA-1.5, ST-Pruner retains 192, 128, or 64 tokens from the original
576-token group and uses $R=72$.
For LLaVA-NeXT, scoring and normalization are performed independently
within each 576-token visual group with $R=72$, and we retain 128, 64, or
32 tokens per group.
For an input with five groups, these settings correspond to at most 640,
320, or 160 retained visual tokens.
For Video-LLaVA, we use $R=32$ independently for each 256-token frame and
retain 17 tokens per frame, giving 136 retained tokens over eight frames.
For Qwen2.5-VL, we use $R=162$ and retain 512, 256, or 128 tokens.

For LLaVA-1.5, LLaVA-NeXT, and Video-LLaVA, randomized low-rank SVD uses
sketch size $q_{\mathrm{svd}}=R$ with two power iterations and no additional
oversampling.
For Qwen2.5-VL, we use $q_{\mathrm{svd}}=R+6$ with two power iterations.
Thus, the default spectral rank is $R=N/8$ within each local visual group
for all evaluated backbones and is not tuned separately for individual
benchmarks or target token budgets.

%

\section{Fine-Grained Hyperparameter Analysis}
\label{sec:supp_hparam}

The main paper summarizes robustness to fusion strategy and spectral rank using mean benchmark-wise retention.
Here we report the complete per-benchmark results and the full hyperparameter sweeps in Tables~\ref{tab:supp_fusion} and \ref{tab:supp_rank}.
For these controlled ablations, retention is normalized by the full-token reproduction from the same evaluation runs:
GQA $61.96$, POPE $85.88$, TextVQA $58.23$, MME$^{\mathrm{P}}$ $1509.76$, and MMB $64.29$.
This matched reference is used only for the design-robustness analysis.

\subsection{Fusion-Strategy Sensitivity}
\label{sec:supp_fusion}

We compare direct multiplicative fusion with sample-adaptive entropy weighting.
For each cue, $H$ denotes the entropy used by the adaptive weighting rule, and the two cue weights are obtained from $\mathrm{softmax}(-H/\tau)$, where the temperature $\tau$ controls the sharpness of the weighting.
Lower-entropy cues therefore receive greater influence.

\begin{table}[!ht]
\centering
\caption{\textbf{Fine-grained fusion-strategy results.}
Avg. is the mean benchmark-wise performance retention relative to the
matched full-token reproduction.}
\label{tab:supp_fusion}
\footnotesize
\renewcommand{\arraystretch}{1.1}
\setlength{\tabcolsep}{12.7pt}
\begin{tabular}{l|ccccc|c}
\toprule
Setting & GQA & POPE & TextVQA & MME$^{\mathrm{P}}$ & MMB & Avg. \\
\midrule
\rowcolor{gray!15}
\multicolumn{7}{c}{\emph{Retain 192 Tokens}} \\
Entropy-weighted ($\tau=0.1$)
& 60.23 & 87.24 & 57.46 & \textbf{1462.73} & 62.24 & \textbf{98.23} \\
Entropy-weighted ($\tau=0.25$)
& 60.23 & 87.22 & 57.42 & 1439.85 & 62.41 & 97.96 \\
Entropy-weighted ($\tau=0.5$)
& 60.33 & \textbf{87.37} & 57.43 & 1430.01 & 62.33 & 97.88 \\
Entropy-weighted ($\tau=1.0$)
& \textbf{60.34} & 87.34 & 57.49 & 1435.92 & 62.24 & 97.95 \\
Entropy-weighted ($\tau=2.0$)
& 60.32 & 87.30 & 57.45 & 1438.44 & 62.24 & 97.95 \\
\rowcolor{oursbg}
\textbf{Direct product (Ours)}
& 60.26 & 87.29 & \textbf{57.59} & 1449.94
& \textbf{62.50} & 98.21 \\
\midrule
\rowcolor{gray!15}
\multicolumn{7}{c}{\emph{Retain 128 Tokens}} \\
Entropy-weighted ($\tau=0.1$)
& 59.11 & 87.33 & \textbf{57.18} & 1409.17 & 61.48 & 96.85 \\
Entropy-weighted ($\tau=0.25$)
& 59.15 & 87.26 & 56.91 & 1422.21 & 61.56 & 96.95 \\
Entropy-weighted ($\tau=0.5$)
& 59.16 & 87.32 & 57.05 & 1430.83 & 61.90 & 97.24 \\
Entropy-weighted ($\tau=1.0$)
& 59.02 & \textbf{87.35} & 56.87 & 1415.81 & 61.90 & 96.94 \\
Entropy-weighted ($\tau=2.0$)
& 59.25 & 87.30 & 57.16 & 1413.78 & 61.65 & 97.00 \\
\rowcolor{oursbg}
\textbf{Direct product (Ours)}
& \textbf{59.38} & 87.26 & 57.06 & \textbf{1432.26}
& \textbf{62.16} & \textbf{97.40} \\
\midrule
\rowcolor{gray!15}
\multicolumn{7}{c}{\emph{Retain 64 Tokens}} \\
Entropy-weighted ($\tau=0.1$)
& 56.47 & 85.59 & 55.37 & 1357.92 & 59.18 & 93.58 \\
Entropy-weighted ($\tau=0.25$)
& 56.75 & 86.45 & \textbf{55.66} & 1361.01 & 58.76 & 93.88 \\
Entropy-weighted ($\tau=0.5$)
& 56.66 & 86.34 & 55.64 & \textbf{1393.65} & 59.35 & 94.43 \\
Entropy-weighted ($\tau=1.0$)
& 56.82 & 86.29 & 55.56 & 1374.63 & 59.69 & 94.30 \\
Entropy-weighted ($\tau=2.0$)
& \textbf{56.85} & 86.26 & 55.41 & 1367.09 & \textbf{59.78} & 94.18 \\
\rowcolor{oursbg}
\textbf{Direct product (Ours)}
& 56.80 & \textbf{86.47} & 55.54 & 1384.42
& \textbf{59.78} & \textbf{94.48} \\
\bottomrule
\end{tabular}
\end{table}

Across the three token budgets, entropy weighting does not provide a
consistent improvement over direct multiplication.
The direct product attains the strongest average retention at the 128-
and 64-token budgets, while the best entropy-weighted setting is only
marginally higher at 192 tokens.
We therefore retain the parameter-free direct product as the default
fusion rule.

\FloatBarrier

\subsection{Spectral-Rank Sensitivity}
\label{sec:supp_rank}

We next fix direct multiplicative fusion and sweep the randomized low-rank SVD dimension $R$.
Performance remains stable over a broad range, with $R=72$ providing the strongest average retention at all three budgets.


\begin{table}[!ht]
\centering
\caption{\textbf{Fine-grained spectral-rank sweep.}
Avg. is the mean benchmark-wise performance retention relative to the
matched full-token reproduction.}
\label{tab:supp_rank}
\footnotesize
\renewcommand{\arraystretch}{0.90}
\setlength{\tabcolsep}{12.7pt}
\begin{tabular}{l|ccccc|c}
\toprule
Rank & GQA & POPE & TextVQA & MME$^{\mathrm{P}}$ & MMB & Avg. \\
\midrule
\rowcolor{gray!15}
\multicolumn{7}{c}{\emph{Retain 192 Tokens}} \\
$R=18$  & 60.15 & 86.31 & 55.03 & 1449.01 & 62.84 & 97.16 \\
$R=36$  & \textbf{60.43} & 87.03 & 57.15 & 1448.93 & \textbf{62.93} & 98.17 \\
\rowcolor{oursbg}
$R=72$ \textbf{(Ours)}
& 60.26 & \textbf{87.29} & \textbf{57.59} & 1449.94 & 62.50 & \textbf{98.21} \\
$R=144$ & 60.25 & 87.07 & 57.17 & \textbf{1455.17} & 62.16 & 97.98 \\
$R=288$ & 60.10 & 87.07 & 56.87 & 1435.42 & 62.59 & 97.70 \\
\midrule
\rowcolor{gray!15}
\multicolumn{7}{c}{\emph{Retain 128 Tokens}} \\
$R=18$  & \textbf{59.69} & 85.31 & 54.23 & \textbf{1442.70} & 61.99 & 96.16 \\
$R=36$  & 59.48 & 86.63 & 56.33 & 1437.65 & 61.90 & 97.02 \\
\rowcolor{oursbg}
$R=72$ \textbf{(Ours)}
& 59.38 & 87.26 & \textbf{57.06} & 1432.26 & 62.16 & \textbf{97.40} \\
$R=144$ & 59.21 & 87.27 & 56.51 & 1430.85 & 61.65 & 96.98 \\
$R=288$ & 59.25 & \textbf{87.45} & 55.81 & 1428.52 & \textbf{62.24} & 96.95 \\
\midrule
\rowcolor{gray!15}
\multicolumn{7}{c}{\emph{Retain 64 Tokens}} \\
$R=18$  & \textbf{57.80} & 82.69 & 52.62 & \textbf{1426.26} & 60.12 & 93.58 \\
$R=36$  & 57.24 & 84.85 & 55.38 & 1386.95 & 59.61 & 94.17 \\
\rowcolor{oursbg}
$R=72$ \textbf{(Ours)}
& 56.80 & 86.47 & \textbf{55.54} & 1384.42 & 59.78 & \textbf{94.48} \\
$R=144$ & 56.88 & 86.41 & 54.40 & 1372.95 & 60.03 & 94.03 \\
$R=288$ & 57.37 & \textbf{86.83} & 53.41 & 1349.80 & \textbf{60.46} & 93.77 \\
\bottomrule
\end{tabular}
\end{table}

Across the complete sweep from $R=18$ to $R=288$, average
retention varies by only $1.05$, $1.24$, and $0.90$ percentage points at
the 192-, 128-, and 64-token budgets, respectively.
The adopted $R=72$ setting achieves the highest average retention at all
three budgets while retaining only one eighth of the 576 available
singular components.
Because the leading randomized-SVD cost scales as $\mathcal{O}(NdR)$ for
a fixed number of subspace iterations, this setting provides a favorable
balance between spectral coverage and scoring cost.

The limited benefit of increasing $R$ is also consistent with useful
structure being concentrated in the leading spectral subspace.
For $R_2>R_1$, extending the retained subspace adds
\[
\bar{s}_i^{(R_2)}-\bar{s}_i^{(R_1)}
=
\sum_{k=R_1+1}^{R_2}\sqrt{\sigma_k}\,U_{ik}^{2}.
\]
Thus, increasing $R$ introduces progressively later singular directions
in addition to the already retained leading components.
The non-monotonic gains in Table~\ref{tab:supp_rank} indicate that these
additional directions provide limited benefit for token selection.

The per-benchmark optima also vary with rank.
At the 64-token budget, $R=18$ performs best on GQA and
MME$^{\mathrm{P}}$, $R=288$ on POPE and MMB, and $R=72$ on
TextVQA and average retention.
Thus, increasing $R$ does not uniformly improve token selection across tasks.

The aggregate results therefore favor a stable middle regime rather than
either an extremely small or unnecessarily large decomposition.
At 192 tokens, $R=36$ and $R=72$ differ by only $0.04$ points in average
retention, while at 128 and 64 tokens the adopted $R=72$ setting remains
the strongest overall.
Moreover, $R=72$ is fixed independently of the target token budget and
corresponds to the same $R=N/8$ rule used throughout our local visual
groups.
Thus, the selected rank reflects a robust operating point rather than
benchmark-specific or budget-specific tuning.

Rank sensitivity is also modest compared with the effect of the token
budget itself.
Across the evaluated ranks, reducing the budget from 192 to 64 tokens
lowers average retention by roughly $3.6$--$4.0$ percentage points,
whereas changing $R$ within a fixed budget changes the average by at most
$1.24$ points.
This suggests that the retained-token budget remains the primary
efficiency--accuracy control, while the spectral rank acts as a secondary
and comparatively robust design parameter.

\begin{figure}[p]
    \centering
    \includegraphics[width=0.96\textwidth]{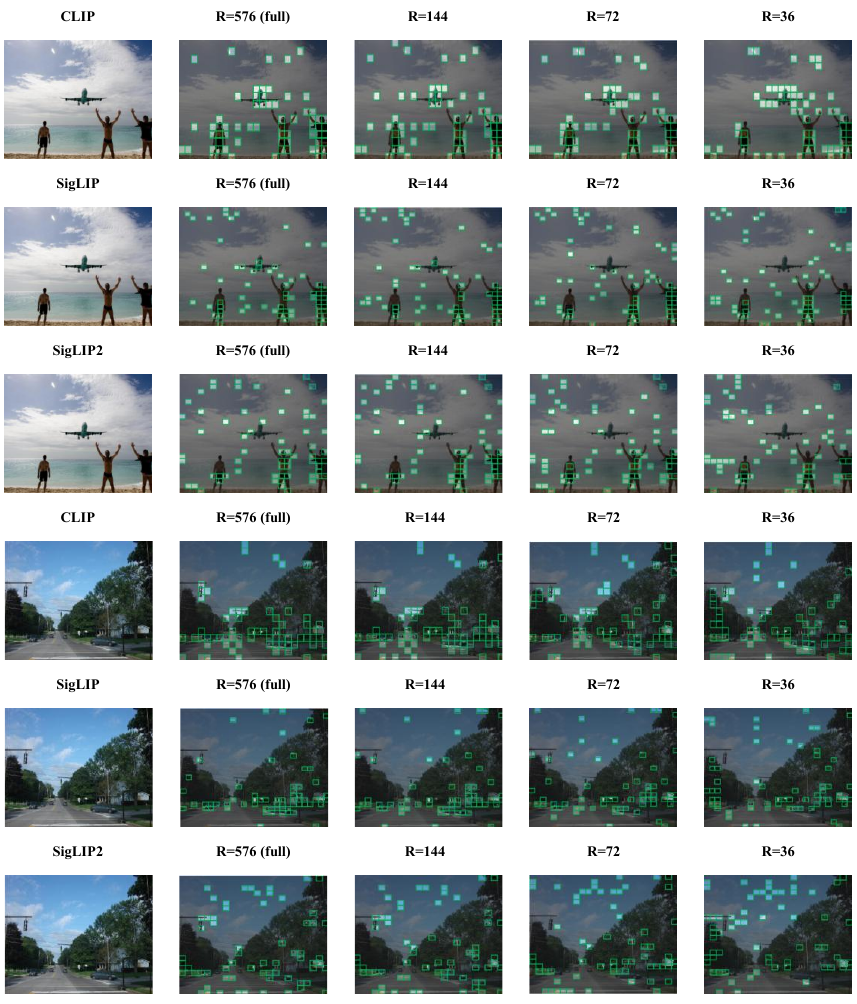}
    \caption{
    \textbf{Qualitative comparison across vision encoders and spectral ranks.}
    Two representative examples are shown.
    For each example, rows correspond to different vision encoders 
    (CLIP, SigLIP, and SigLIP2), and columns show the original image and
    top-$K$ selections obtained with full-rank scoring and reduced-rank
    approximations ($R=144, 72, 36$).
    Across encoder families and rank settings, the selections remain broadly
    concentrated on similar salient object and scene regions, with $R=72$
    showing patterns close to full-rank scoring.}
    \label{fig:supp_encoder_rank}
\end{figure}

Figure~\ref{fig:supp_encoder_rank} complements this quantitative result by
visualizing top-64 selections from several vision encoders.
Although the encoders use different feature spaces and tokenizations,
their reduced-rank selections remain broadly concentrated on similar
salient object and scene regions and remain qualitatively close to
full-rank scoring.

\FloatBarrier


\section{Additional Spectral-Design Ablations}
\label{sec:supp_design}

\subsection{Feature-Space Ablation}
\label{sec:supp_feature_space}

The default ST-Pruner computes spectral saliency from the
pre-projector ViT features.
We compare this choice with computing the same spectral score from the
4096-dimensional projector output while keeping all other settings
fixed, including $\sqrt{\sigma}$ weighting, $R=72$, randomized
low-rank SVD, and direct multiplicative fusion.

\FloatBarrier

\begin{table}[!ht]
\centering
\caption{\textbf{Feature-space ablation on LLaVA-1.5-7B.}
We compare spectral scoring on 1024-dimensional pre-projector ViT
features and 4096-dimensional projector outputs while keeping all other
settings fixed.}
\label{tab:supp_feature_space}
\footnotesize
\renewcommand{\arraystretch}{1.05}
\setlength{\tabcolsep}{13.7pt}
\begin{tabular}{l|ccccc}
\toprule
Feature space & POPE & GQA & TextVQA & MME$^{\mathrm{P}}$ & MMB \\
\midrule
\rowcolor{gray!15}
\multicolumn{6}{c}{\emph{Retain 192 Tokens}} \\
\rowcolor{oursbg}
\textbf{ViT-1024 (Ours)}
& \textbf{87.29} & 60.26 & \textbf{57.59} & 1449.94 & 62.50 \\
Projector-4096
& 87.02 & \textbf{60.42} & 57.55 & \textbf{1452.84} & \textbf{62.93} \\
\midrule
\rowcolor{gray!15}
\multicolumn{6}{c}{\emph{Retain 128 Tokens}} \\
\rowcolor{oursbg}
\textbf{ViT-1024 (Ours)}
& \textbf{87.26} & 59.38 & \textbf{57.06} & 1432.26 & \textbf{62.16} \\
Projector-4096
& 87.13 & \textbf{59.54} & 56.98 & \textbf{1437.73} & 61.99 \\
\midrule
\rowcolor{gray!15}
\multicolumn{6}{c}{\emph{Retain 64 Tokens}} \\
\rowcolor{oursbg}
\textbf{ViT-1024 (Ours)}
& \textbf{86.47} & 56.80 & 55.54 & 1384.42 & 59.78 \\
Projector-4096
& 85.87 & \textbf{57.17} & \textbf{55.74} & \textbf{1388.24}
& \textbf{60.80} \\
\bottomrule
\end{tabular}
\end{table}

The two feature spaces yield comparable performance across all three
token budgets.
Because the leading randomized-SVD complexity is
$\mathcal{O}(NdR)$, using the 1024-dimensional ViT representation
reduces the dominant SVD computation by a factor of four relative to
the 4096-dimensional projector representation for the same $N$ and
$R$.
We therefore use the pre-projector ViT features as the default
feature space for spectral scoring, providing a more favorable
accuracy--computation trade-off.
The comparable performance also suggests that useful low-rank structure
is already accessible in the vision-encoder representation, without
requiring the higher-dimensional projector representation for spectral scoring.

\FloatBarrier

\subsection{Singular-Value Weighting}
\label{sec:supp_weighting}

To isolate the role of singular-value weighting, we keep the squared
token loading $U_{ik}^{2}$ fixed and vary only the exponent $p$ in
\[
\bar{s}_i^{(p)}
=
\sum_{k=1}^{R}\sigma_k^{p}U_{ik}^{2}.
\]
This sweep separates four regimes: uniform weighting over the retained
components ($p=0$), the proposed square-root weighting ($p=0.5$), direct
singular-value weighting ($p=1$), and deliberately aggressive quadratic
weighting ($p=2$).
All other settings are fixed to the default ViT feature space, $R=72$,
randomized low-rank SVD, and direct multiplicative fusion.

\FloatBarrier

\begin{table}[!ht]
\centering
\caption{\textbf{Singular-value weighting ablation on LLaVA-1.5-7B.}
The exponent $p$ controls the strength of singular-value weighting in
$\sum_{k\leq R}\sigma_k^p U_{ik}^2$.
Avg. is the mean benchmark-wise performance retention relative to the
matched full-token reproduction.}
\label{tab:supp_weighting}
\footnotesize
\renewcommand{\arraystretch}{1.05}
\setlength{\tabcolsep}{13.7pt}
\begin{tabular}{l|ccccc|c}
\toprule
Weighting & GQA & POPE & TextVQA & MME$^{\mathrm{P}}$ & MMB & Avg. \\
\midrule
\rowcolor{gray!15}
\multicolumn{7}{c}{\emph{Retain 192 Tokens}} \\
$p=0$ (uniform)
& 60.28 & 87.22 & 57.32 & 1446.14 & 62.41 & 98.03 \\
\rowcolor{oursbg}
$p=0.5$ \textbf{(Ours)}
& 60.26 & \textbf{87.29} & \textbf{57.59} & 1449.94
& 62.50 & \textbf{98.21} \\
$p=1$ (linear)
& \textbf{60.34} & 87.25 & 57.23 & \textbf{1450.04}
& \textbf{62.67} & 98.16 \\
$p=2$ (quadratic)
& 56.61 & 82.98 & 49.53 & 1269.44 & 51.79 & 87.54 \\
\midrule
\rowcolor{gray!15}
\multicolumn{7}{c}{\emph{Retain 128 Tokens}} \\
$p=0$ (uniform)
& 59.06 & \textbf{87.39} & \textbf{57.09} & 1425.92 & 61.90 & 97.17 \\
\rowcolor{oursbg}
$p=0.5$ \textbf{(Ours)}
& 59.38 & 87.26 & 57.06 & \textbf{1432.26}
& \textbf{62.16} & \textbf{97.40} \\
$p=1$ (linear)
& \textbf{59.49} & 87.04 & 56.63 & 1431.68 & 61.73 & 97.09 \\
$p=2$ (quadratic)
& 51.17 & 73.45 & 46.97 & 1103.77 & 38.01 & 76.20 \\
\midrule
\rowcolor{gray!15}
\multicolumn{7}{c}{\emph{Retain 64 Tokens}} \\
$p=0$ (uniform)
& 56.77 & 86.43 & 55.09 & 1367.64 & 59.69 & 94.06 \\
\rowcolor{oursbg}
$p=0.5$ \textbf{(Ours)}
& 56.80 & \textbf{86.47} & 55.54 & 1384.42
& \textbf{59.78} & \textbf{94.48} \\
$p=1$ (linear)
& \textbf{57.18} & 85.27 & \textbf{55.55} & \textbf{1391.47}
& 59.61 & 94.37 \\
$p=2$ (quadratic)
& 43.66 & 40.97 & 43.66 & 847.45 & 29.25 & 58.96 \\
\bottomrule
\end{tabular}
\end{table}

The proposed $p=0.5$ weighting gives the highest average retention at
all three token budgets.
The $p=0$ variant is nevertheless competitive.
Importantly, this result should not be interpreted as showing that the
spectral decomposition or singular-value ordering is unnecessary.
At $p=0$, the score reduces to
\[
\bar{s}_i^{(0)}
=
\sum_{k=1}^{R}U_{ik}^{2}
=
\lVert \mathbf{U}_R(i,:)\rVert_2^2,
\]
which measures the participation of token $i$ in the retained leading
$R$-dimensional left-singular subspace.
Because $R<N$, these row norms remain token-dependent, while the
truncation itself has already selected the components associated with
the leading singular values.
For the present $N\leq d$ setting, this differs from the full-rank case
$R=N$: the resulting square $\mathbf{U}$ is orthogonal, so
$\mathbf{U}\mathbf{U}^{\top}=\mathbf{I}_N$, and therefore
$\sum_{k=1}^{N}U_{ik}^{2}=1$ for every token.
Thus, the strong $p=0$ result indicates that the truncated dominant
subspace is itself informative rather than that singular-spectrum
structure is absent.

This distinction also clarifies the role of truncation itself.
The quantity
$\lVert \mathbf{U}_R(i,:)\rVert_2^2$ has the form of a rank-$R$
leverage score: it measures how strongly token $i$ participates in the
dominant left-singular subspace without differentiating the retained
components by magnitude.
The usefulness of $p=0$ therefore comes from restricting the score to the
leading subspace selected by the singular-value ordering.
If all $N$ left-singular directions were retained with uniform weighting,
the row-norm identity above would remove this token-wise discrimination.
Our $p=0.5$ score can thus be viewed as refining an already informative
truncated-subspace signal by reintroducing component strength in a
tempered form.

Relative to $p=0$, the proposed $p=0.5$ improves average retention by
$0.18$, $0.23$, and $0.42$ percentage points at 192, 128, and 64 tokens,
respectively.
Direct linear weighting ($p=1$) remains close, but its average retention
is lower than $p=0.5$ at every budget.
Increasing the exponent further to $p=2$ causes a sharp degradation,
with average retention falling to $87.54\%$, $76.20\%$, and $58.96\%$
as the token budget decreases.
These results show a consistent peak at the square-root setting:
$p=0.5$ preserves singular-value strength while tempering concentration
on the strongest components, providing a moderate balance between
uniform subspace participation and increasingly aggressive
magnitude-based weighting.

The benefit of this balance becomes slightly more visible as compression
becomes stronger.
Relative to uniform weighting, the average-retention gain of $p=0.5$
increases from $0.18$ points at 192 tokens to $0.42$ points at 64 tokens,
while its advantage over $p=2$ widens from $10.67$ to $35.52$ points.
This widening gap suggests that excessively weighting the strongest
components becomes particularly harmful under aggressive token compression.

\FloatBarrier

\renewcommand{\bottomfraction}{0.55}
\renewcommand{\topfraction}{0.90}
\renewcommand{\textfraction}{0.05}

\subsection{Qualitative Spectral Analysis}
\label{sec:supp_qualitative}

To inspect how different parts of the singular spectrum contribute to
the spectral score, we first compute a full SVD of each 576-token
feature matrix and partition the ordered singular components into
disjoint component-index bands.
For each band $\mathcal{B}$, we apply the same band-restricted score,
\[
\bar{s}^{(\mathcal{B})}_i
=
\sum_{k\in\mathcal{B}}\sqrt{\sigma_k}\,U_{ik}^{2},
\]
and visualize its top-$K$ patches.
This analysis is diagnostic only; the deployed method directly computes
a randomized truncated SVD.
Accordingly, the leading $1$--$72$ band is the closest full-SVD
analogue to our default $R=72$ approximation, although the two
decompositions are not numerically identical.

\begin{figure}[!b]
    \centering
    \includegraphics[
        width=0.95\textwidth,
        height=0.68\textheight,
        keepaspectratio
    ]{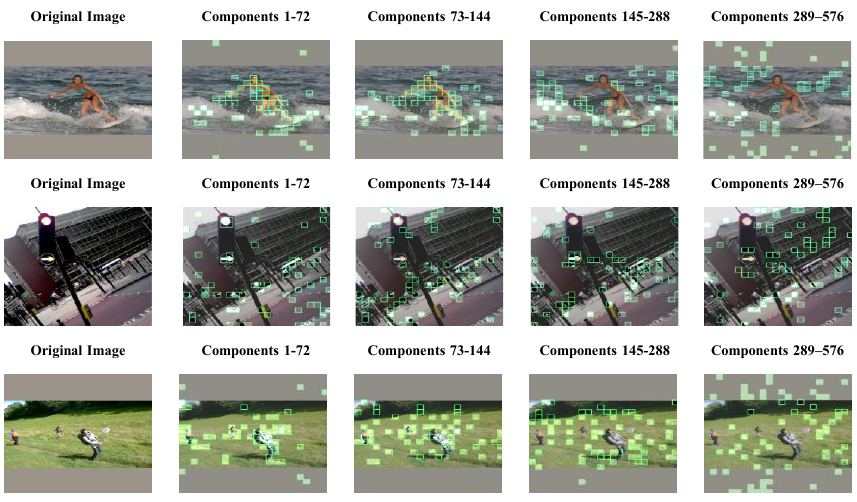}
    \caption{\textbf{Spectral hierarchy on general visual content.}
    Three examples span different scene structures.
    Columns show the original image and top-$K$ selections from component
    bands $1$--$72$, $73$--$144$, $145$--$288$, and $289$--$576$.
    Leading bands emphasize coherent foreground structure, whereas
    lower-energy bands become increasingly dispersed.}
    \label{fig:supp_spectral_hierarchy_general}
\end{figure}

Figure~\ref{fig:supp_spectral_hierarchy_general} presents three general
visual examples spanning different scene structures, from a relatively
simple person-centric case to more cluttered object and background
configurations.
Across these examples, the leading $1$--$72$ band emphasizes coherent
foreground content and visually salient structures.
The $73$--$144$ band preserves a similar focus on coherent foreground
and salient visual structure, whereas the $145$--$288$ and
$289$--$576$ bands become increasingly dispersed over peripheral and
background regions.
This progression provides qualitative evidence that the leading
singular components contain more useful structural information for
token selection, consistent with the limited benefit of increasing
$R$ in Table~\ref{tab:supp_rank} and Figure~\ref{fig:supp_encoder_rank}.

\begin{figure}[!htb]
    \centering
    \includegraphics[
        width=0.95\textwidth,
        height=0.68\textheight,
        keepaspectratio
    ]{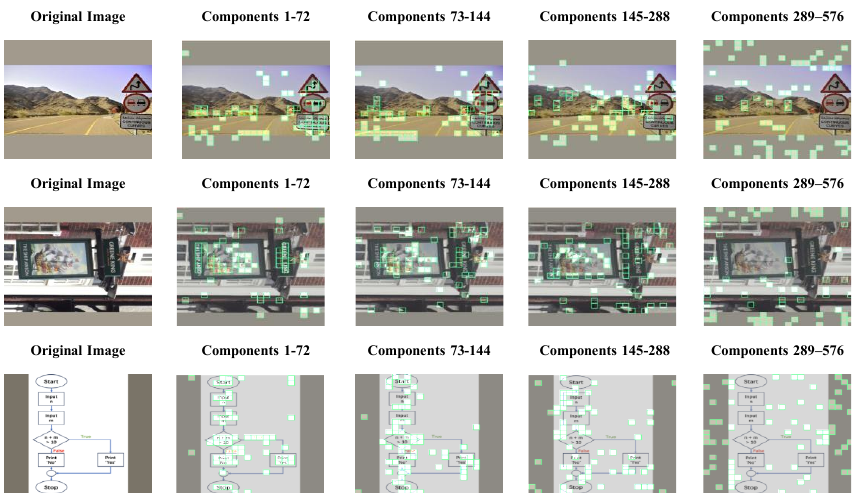}
    \caption{\textbf{Spectral hierarchy on text-bearing visual content.}
    Three OCR-relevant examples use the same component bands as
    Figure~\ref{fig:supp_spectral_hierarchy_general}.
    Leading bands preserve localized text-bearing regions, whereas later
    bands become increasingly dispersed.}
    \label{fig:supp_spectral_hierarchy_ocr}
\end{figure}

Figure~\ref{fig:supp_spectral_hierarchy_ocr} focuses on three
text-bearing examples in which OCR-relevant local detail is important.
The leading bands preserve localized regions containing traffic signs,
screen text, and labels inside a diagram, while later bands shift toward
more diffuse low-energy detail.
Together, the two visualizations indicate that dominant spectral
components capture not only object-centric foreground structure but also
fine-grained text-bearing regions, consistent with the strong TextVQA
retention observed in our quantitative results.

\FloatBarrier

\section{Complete Efficiency Sweep}
\label{sec:supp_efficiency}

The main paper reports the matched 320-token comparison.
Here we provide the complete 640, 320, and 160-token sweep measured
over the full POPE evaluation set of 8,910 samples on a single NVIDIA
V100 GPU under FP16.
The unpruned LLaVA-NeXT-7B baseline uses 2880 visual tokens and requires
34.92T analytical LLM prefill FLOPs, 679.2 ms prefill latency,
48.8 ms decode latency, 727.9 ms total generation latency, and
1542 MB of visual-prefix KV cache while reaching 86.5 POPE F1.
Generation uses a maximum of 16 new tokens.

All reported prefill and total latencies include each method's pruning
or scoring overhead inside the timed generation call.
For ST-Pruner, this includes randomized low-rank SVD, CLIP text
encoding, patch--text relevance computation, multiplicative fusion,
and top-$K$ selection.
No pruning scores are precomputed.

\begin{table}[!htb]
\centering
\caption{\textbf{Complete inference-efficiency sweep on LLaVA-NeXT-7B.}
Values in parentheses denote speedup or FLOPs reduction factors relative
to the 2880-token baseline.
All latencies are measured over the full POPE set.
For SparseVLM, FLOPs and KV cache use its actual layer-wise token
trajectory rather than the final retained token count.}
\label{tab:supp_efficiency_full}
\scriptsize
\renewcommand{\arraystretch}{1.1}
\setlength{\tabcolsep}{10.7pt}
\begin{tabular}{l|c|c|c|c|c|c|c}
\toprule
Method & Tokens & FLOPs (T) & Prefill (ms) & Decode (ms) &
Total (ms) & KV (MB) & F1 \\
\midrule
LLaVA-NeXT-7B
& 2880 & 34.92 & 679.2 & 48.8 & 727.9 & 1542 & 86.5 \\

\rowcolor{gray!15}
\multicolumn{8}{c}{\emph{Retain 640 Tokens}} \\
SparseVLM \venue{ICML'25}
& 640\rlap{$^\dagger$} & 11.05 {\tiny($\times3.2$)}
& 368.4 {\tiny($\times1.8$)}
& 44.8 {\tiny($\times1.1$)}
& 413.1 {\tiny($\times1.8$)}
& 505 & 85.3 \\
VisionZip \venue{CVPR'25}
& 640 & 8.33 {\tiny($\times4.2$)}
& \textbf{222.7} {\tiny($\times3.1$)}
& 34.5 {\tiny($\times1.4$)}
& \textbf{257.2} {\tiny($\times2.8$)}
& 368 & 86.0 \\
DivPrune \venue{CVPR'25}
& 640 & 8.33 {\tiny($\times4.2$)}
& 286.5 {\tiny($\times2.4$)}
& 32.5 {\tiny($\times1.5$)}
& 319.0 {\tiny($\times2.3$)}
& 368 & 86.0 \\
VisPruner \venue{ICCV'25}
& 640 & 8.33 {\tiny($\times4.2$)}
& 267.9 {\tiny($\times2.5$)}
& 36.8 {\tiny($\times1.3$)}
& 304.7 {\tiny($\times2.4$)}
& 368 & 85.9 \\
CDPruner \venue{NeurIPS'25}
& 640 & 8.33 {\tiny($\times4.2$)}
& 290.5 {\tiny($\times2.3$)}
& 34.5 {\tiny($\times1.4$)}
& 324.9 {\tiny($\times2.2$)}
& 368 & 86.5 \\
\rowcolor{spectralbg}
\textbf{ST-Pruner (Spectral)}
& 640 & 8.33 {\tiny($\times4.2$)}
& 248.7 {\tiny($\times2.7$)}
& \textbf{31.9} {\tiny($\times1.5$)}
& 280.6 {\tiny($\times2.6$)}
& 368 & 86.3 \\
\rowcolor{oursbg}
\textbf{ST-Pruner (Ours)}
& 640 & 8.33 {\tiny($\times4.2$)}
& 272.1 {\tiny($\times2.5$)}
& 37.8 {\tiny($\times1.3$)}
& 309.9 {\tiny($\times2.3$)}
& 368 & \textbf{87.0} \\

\rowcolor{gray!15}
\multicolumn{8}{c}{\emph{Retain 320 Tokens}} \\
SparseVLM \venue{ICML'25}
& 320\rlap{$^\dagger$} & 7.81 {\tiny($\times4.5$)}
& 299.3 {\tiny($\times2.3$)}
& 44.1 {\tiny($\times1.1$)}
& 343.4 {\tiny($\times2.1$)}
& 359 & 76.9 \\
VisionZip \venue{CVPR'25}
& 320 & 4.53 {\tiny($\times7.7$)}
& \textbf{142.5} {\tiny($\times4.8$)}
& 36.0 {\tiny($\times1.4$)}
& \textbf{178.4} {\tiny($\times4.1$)}
& 200 & 82.3 \\
DivPrune \venue{CVPR'25}
& 320 & 4.53 {\tiny($\times7.7$)}
& 187.5 {\tiny($\times3.6$)}
& 33.6 {\tiny($\times1.5$)}
& 221.1 {\tiny($\times3.3$)}
& 200 & 83.4 \\
VisPruner \venue{ICCV'25}
& 320 & 4.53 {\tiny($\times7.7$)}
& 163.0 {\tiny($\times4.2$)}
& 34.9 {\tiny($\times1.4$)}
& 197.9 {\tiny($\times3.7$)}
& 200 & 81.3 \\
CDPruner \venue{NeurIPS'25}
& 320 & 4.53 {\tiny($\times7.7$)}
& 177.7 {\tiny($\times3.8$)}
& 34.2 {\tiny($\times1.4$)}
& 211.9 {\tiny($\times3.4$)}
& 200 & 85.8 \\
\rowcolor{spectralbg}
\textbf{ST-Pruner (Spectral)}
& 320 & 4.53 {\tiny($\times7.7$)}
& 165.4 {\tiny($\times4.1$)}
& \textbf{31.8} {\tiny($\times1.5$)}
& 197.2 {\tiny($\times3.7$)}
& 200 & 83.1 \\
\rowcolor{oursbg}
\textbf{ST-Pruner (Ours)}
& 320 & 4.53 {\tiny($\times7.7$)}
& 188.0 {\tiny($\times3.6$)}
& 36.0 {\tiny($\times1.4$)}
& 224.0 {\tiny($\times3.2$)}
& 200 & \textbf{86.1} \\

\rowcolor{gray!15}
\multicolumn{8}{c}{\emph{Retain 160 Tokens}} \\
SparseVLM \venue{ICML'25}
& 160\rlap{$^\dagger$} & 5.87 {\tiny($\times5.9$)}
& 249.6 {\tiny($\times2.7$)}
& 43.5 {\tiny($\times1.1$)}
& 293.1 {\tiny($\times2.5$)}
& 270 & -- \\
VisionZip \venue{CVPR'25}
& 160 & 2.64 {\tiny($\times13.3$)}
& \textbf{120.8} {\tiny($\times5.6$)}
& 35.0 {\tiny($\times1.4$)}
& \textbf{155.8} {\tiny($\times4.7$)}
& 116 & 74.9 \\
DivPrune \venue{CVPR'25}
& 160 & 2.64 {\tiny($\times13.3$)}
& 135.6 {\tiny($\times5.0$)}
& \textbf{32.2} {\tiny($\times1.5$)}
& 167.8 {\tiny($\times4.3$)}
& 116 & 79.6 \\
VisPruner \venue{ICCV'25}
& 160 & 2.64 {\tiny($\times13.3$)}
& 130.2 {\tiny($\times5.2$)}
& 32.4 {\tiny($\times1.5$)}
& 162.6 {\tiny($\times4.5$)}
& 116 & 75.4 \\
CDPruner \venue{NeurIPS'25}
& 160 & 2.64 {\tiny($\times13.3$)}
& 140.3 {\tiny($\times4.8$)}
& 34.2 {\tiny($\times1.4$)}
& 174.5 {\tiny($\times4.2$)}
& 116 & 84.4 \\
\rowcolor{spectralbg}
\textbf{ST-Pruner (Spectral)}
& 160 & 2.64 {\tiny($\times13.3$)}
& 143.7 {\tiny($\times4.7$)}
& 32.8 {\tiny($\times1.5$)}
& 176.5 {\tiny($\times4.1$)}
& 116 & 76.6 \\
\rowcolor{oursbg}
\textbf{ST-Pruner (Ours)}
& 160 & 2.64 {\tiny($\times13.3$)}
& 164.0 {\tiny($\times4.1$)}
& 33.1 {\tiny($\times1.5$)}
& 197.1 {\tiny($\times3.7$)}
& 116 & \textbf{85.0} \\
\bottomrule
\end{tabular}

\vspace{2pt}
{\scriptsize
$^\dagger$SparseVLM progressively prunes inside the LLM.
Its FLOPs and KV cache are therefore computed from the layer-wise token
trajectory rather than from the final token count.}
\end{table}

\FloatBarrier

\paragraph{Layer-wise accounting for SparseVLM.}
SparseVLM does not enter every decoder layer with its final retained
token count.
For the 640, 320, and 160-token targets, respectively, the measured
visual-token trajectories are
$2880\!\rightarrow\!1400\!\rightarrow\!900\!\rightarrow\!640$,
$2880\!\rightarrow\!1100\!\rightarrow\!600\!\rightarrow\!320$, and
$2880\!\rightarrow\!800\!\rightarrow\!400\!\rightarrow\!160$
after the corresponding progressive pruning stages.
We therefore sum analytical LLM FLOPs and KV-cache usage according to
the sequence length entering each layer.
This gives 11.05T, 7.81T, and 5.87T prefill FLOPs for the three target
budgets, rather than assigning SparseVLM the cost of a pre-LLM method
that starts every layer at the final token count.

\paragraph{Cost of the semantic branch.}
The spectral-only variant isolates the cost of the spectral branch.
Moving from spectral-only to the complete ST-Pruner adds approximately
23.4, 22.6, and 20.3 ms of prefill latency at 640, 320, and 160 tokens,
respectively.
The corresponding POPE F1 changes from
$86.3$ to $87.0$, $83.1$ to $86.1$, and $76.6$ to $85.0$.
Thus, the additional semantic scoring cost remains modest while its
accuracy contribution becomes increasingly important at tighter token
budgets.


\FloatBarrier

\end{document}